\documentclass[10pt,twocolumn,letterpaper]{article}

\usepackage{iccv}
\usepackage{times}
\usepackage{epsfig}
\usepackage{graphicx}
\usepackage{amsmath}
\usepackage{amssymb}

\usepackage{setspace}
\usepackage{xcolor}
\usepackage{subcaption}
\usepackage{wrapfig}
\usepackage{pbox}
\usepackage{adjustbox}
\usepackage{floatrow}
\usepackage{subcaption}

\usepackage[pagebackref=true,breaklinks=true,letterpaper=true,colorlinks,bookmarks=false]{hyperref}

\usepackage{amsmath,amsfonts,bm}

\def\eqref#1{equation~\ref{#1}}

\def\ceil#1{\lceil #1 \rceil}

\def\1{\bm{1}}

\DeclareMathAlphabet{\mathsfit}{\encodingdefault}{\sfdefault}{m}{sl}
\SetMathAlphabet{\mathsfit}{bold}{\encodingdefault}{\sfdefault}{bx}{n}

\newcommand{\minisection}[1]{\textbf{#1}\hspace{0.3em}}

\definecolor{calblue}{RGB}{52, 116, 180}
\definecolor{calgold}{RGB}{252, 200, 57}
\definecolor{mixgreen}{RGB}{82, 158, 119}

\newenvironment{smatrix}
  {\left[\begin{smallmatrix}}
  {\end{smallmatrix}\right]}

\iccvfinalcopy 

\ificcvfinal\fi
\begin{document}

\title{Blurring the Line Between Structure and Learning\\to Optimize and Adapt Receptive Fields}

\author{
Evan Shelhamer \quad Dequan Wang \quad Trevor Darrell\\
UC Berkeley\\
{\tt\small {shelhamer,dqwang,trevor}@cs.berkeley.edu}
}

\maketitle

\begin{abstract}
The visual world is vast and varied, but its variations divide into structured and unstructured factors.
We compose free-form filters and structured Gaussian filters, optimized end-to-end, to factorize deep representations and learn both local features and their degree of locality.
Our semi-structured composition is strictly more expressive than free-form filtering, and changes in its structured parameters would require changes in free-form architecture.
In effect this optimizes over receptive field size and shape, tuning locality to the data and task.
Dynamic inference, in which the Gaussian structure varies with the input, adapts receptive field size to compensate for local scale variation.
Optimizing receptive field size improves semantic segmentation accuracy on Cityscapes by 1-2 points for strong dilated and skip architectures and by up to 10 points for suboptimal designs.
Adapting receptive fields by dynamic Gaussian structure further improves results, equaling the accuracy of free-form deformation while improving efficiency.
\end{abstract}

\section{Introduction}

Although the visual world is varied, it nevertheless has ubiquitous structure.
Structured factors, such as scale, admit clear theories and efficient representation design.
Unstructured factors, such as what makes a cat look like a cat, are too complicated to model analytically, requiring free-form representation learning.
How can recognition harness structure without restraining the representation?

Free-form representations are structure-agnostic, making them general, but not exploiting structure is computationally and statistically inefficient.
Structured representations like steerable filtering \cite{freeman1991design,simoncelli1995steerable,jacobsen2016structured}, scattering \cite{bruna2013invariant,sifre2013rotation}, and steerable networks \cite{cohen2017steerable} are efficient but constrained to the chosen structures.
We propose a new, semi-structured compositional filtering approach to blur the line between free-form and structured representations and learn both.
Doing so learns local features and the degree of locality.

\begin{figure}[htp]
\centering
\includegraphics[width=0.95\linewidth]{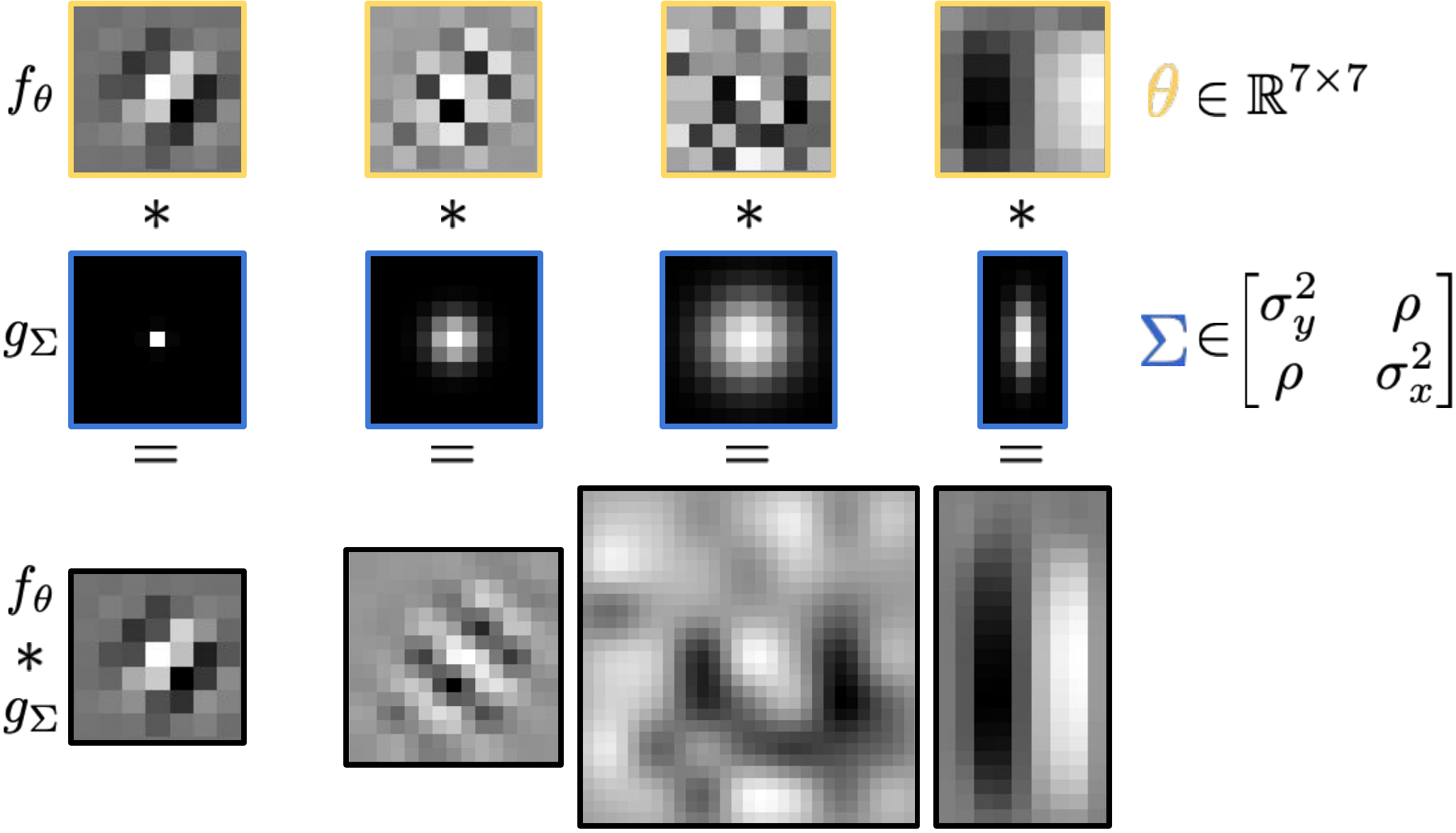}
\vspace{-1mm}
\caption{
We compose free-form filters $f_\theta$ and structured Gaussian filters $g_\Sigma$ by convolution $*$ to define a more general family of semi-structured filters than can be learned by either alone.
Our composition makes receptive field scale, aspect, and orientation differentiable in a low-dimensional parameterization for efficient end-to-end learning.
}
\label{fig:composition}
\end{figure}

Free-form filters, directly defined by the parameters, are general and able to cope with unknown variations, but are parameter inefficient.
Structured factors, such as scale and orientation, are enumerated like any other variation, and require duplicated learning across different layers and channels.
Nonetheless, end-to-end learning of free-form parameters is commonly the most accurate approach to complex visual recognition tasks when there is sufficient data.

Structured filters, indirectly defined as a function of the parameters, are theoretically clear and parameter efficient, but constrained.
Their effectiveness hinges on whether or not they encompass the true structure of the data.
If not, the representation is limiting, and subject to error.
At least, this is a danger when \emph{substituting} structure to replace learning.

\begin{figure}[t]
\centering
\includegraphics[width=0.95\linewidth,page=1]{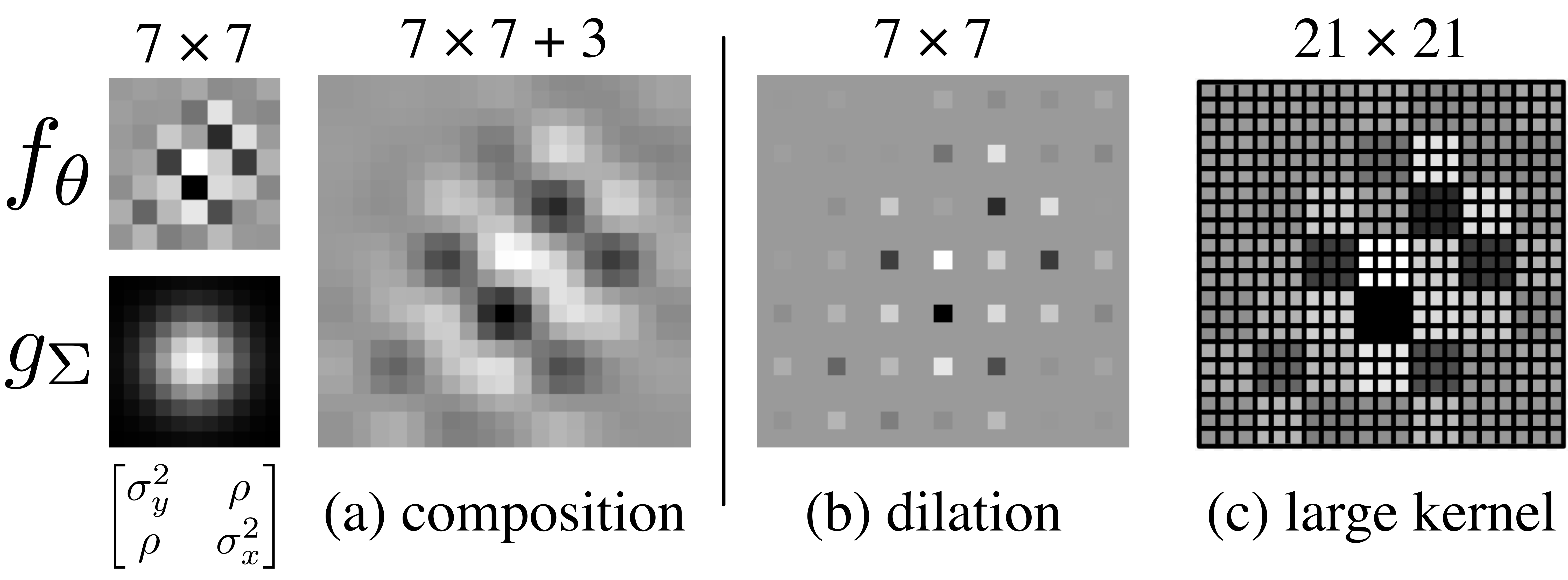}
\vspace{-2mm}
\caption{
Our composition (a) cannot be practically reduced to dilation (b) or more free-form parameters (c).
Adding more parameters, here about $10\times$ by $21^2$ vs. $7^2$, is inefficient and limited: more parameters take more data and optimization to learn and the maximum scale is fixed.
Dilating the filter has side effects and is constrained: sparse sampling causes artifacts and the sparsity and scale are fixed.
}
\label{fig:competition}
\end{figure}

We \emph{compose} free-form and structured filters, as shown in Figure \ref{fig:composition}, and learn both end-to-end.
Free-form filters are not constrained by our composition.
This makes our approach more expressive, not less, while still able to efficiently learn the chosen structured factors.
In this way our semi-structured networks can reduce to existing networks as a special case.
At the same time, our composition can learn different receptive fields that cannot be realized in the standard parameterization of free-form filters.
Adding more free-form parameters or dilating cannot learn the same family of filters.
Figure \ref{fig:competition} offers one example of the impracticality of architectural alternatives.

Gaussian structure represents scale, aspect, and orientation through covariance \cite{lindeberg1994scale-space}.
Optimizing these factors carries out a form of differentiable architecture search over receptive fields, reducing the need for onerous hand-design or expensive discrete search.
Any 2D Gaussian has the same, low number of covariance parameters no matter its spatial extent, so receptive field optimization is low-dimensional and efficient.
Because the Gaussian is smooth, our filtering is guaranteed to be proper from a signal processing perspective and avoid aliasing.

Our contributions include:
(1) defining semi-structured compositional filtering to bridge classic ideas for scale-space representation design and current practices for representation learning,
(2) exploring a variety of receptive fields that our approach can learn,
and
(3) adapting receptive fields with accurate and efficient dynamic Gaussian structure.

\section{Related Work}

Composing structured Gaussian filters with free-form learned filters draws on structured filter design and representation learning.
Our work is inspired by the transformation invariance of scale-space \cite{lindeberg1994scale-space}, the parsimony of steerable filtering \cite{freeman1991design,perona1995deformable,bruna2013invariant,cohen2017steerable}, and the adaptivity of dynamic inference \cite{olshausen1993neurobiological,jaderberg2015spatial,de-brabandere2016dynamic,dai2017deformable}.
Analysis that the effective receptive field size of deep networks is limited \cite{luo2016understanding}, and only is a fraction of the theoretical size, motivates our goal of making unbounded receptive field size and varied receptive field shapes practically learnable.

\minisection{Transformation Invariance}
Gaussian scale-space and its affine extension connect covariance to spatial structure for transformation invariance \cite{lindeberg1994scale-space}.
We jointly learn structured transformations via Gaussian covariance and features via free-form filtering.
Enumerative methods cover a set of transformations, rather than learning to select transformations:
image pyramids \cite{burt1983laplacian} and feature pyramids \cite{kanazawa2014locally,shelhamer2016fully,lin2017feature} cover scale,
scattering \cite{bruna2013invariant} covers scales and rotations, and
steerable networks \cite{cohen2017steerable} cover discrete groups.
Our learning and inferring covariance relates to scale selection \cite{lindeberg1998feature}, as exemplified by the scale invariant feature transform \cite{lowe2004distinctive}.
Scale-adaptive convolution \cite{zhang2017scale} likewise selects scales, but without our Gaussian structure and smoothness.

\minisection{Steering}
Steering indexes a continuous family of filters by linearly weighting a structured basis, such as Gaussian derivatives.
Steerable filters \cite{freeman1991design} index orientation and deformable kernels \cite{perona1995deformable} index orientation and scale.
Such filters can be stacked into a deep, structured network \cite{jacobsen2016structured}.
These methods have elegant structure, but are constrained to it.
We make use of Gaussian structure, but keep generality by composing with free-form filters.

\minisection{Dynamic Inference}
Dynamic inference adapts the model to each input.
Dynamic routing \cite{olshausen1993neurobiological},
spatial transformers \cite{jaderberg2015spatial},
dynamic filter nets \cite{de-brabandere2016dynamic}, and
deformable convolution \cite{dai2017deformable}
are all dynamic, but lack local structure.
We incorporate Gaussian structure to improve efficiency while preserving accuracy.

Proper signal processing, by blurring when downsampling, improves the shift-equivariance of learned filtering \cite{zhang2019making}.
We reinforce these results with our experiments on blurred dilation, to complement their focus on blurred stride.
While we likewise blur, and confirm the need for smoothing to prevent aliasing, our focus is on how to jointly learn and compose structured and free-form filters.


\section{A Clear Review of Blurring}
\label{sec:blur}

We introduce the elements of our chosen structured filters first, and then compose free-form filters with this structure in the next section.
While the Gaussian and scale-space ideas here are classic, our end-to-end optimized composition and its use for receptive field learning are novel. 

\subsection{Gaussian Structure}

The choice of structure determines the filter characteristics that can be represented and learned.

We choose Gaussian structure.
For modeling, it is differentiable for end-to-end learning, low-dimensional for efficient optimization, and still expressive enough to represent a variety of shapes.
For signal processing, it is smooth and admits efficient filtering.
In particular, the Gaussian has these attractive properties for our purposes:
\begin{itemize}
\setlength\itemsep{0.25em}
\item shift-invariance for convolutional filtering,
\item normalization to preserve input and gradient norms for stable optimization,  
\item separability to reduce computation by replacing a 2D filter with two 1D filters,
\item and cascade smoothing from semi-group structure to decompose filtering into smaller, cumulative steps.
\end{itemize}
In fact, the Gaussian is the unique filter satisfying these and further scale-space axioms \cite{koenderink1984structure,babaud1986uniqueness,lindeberg1994scale-space}.

The Gaussian kernel in 2-D is
\begin{equation}
G(\mathbf{x}; \boldsymbol\Sigma) = \frac{1}{2\pi\sqrt{\operatorname{det}\boldsymbol\Sigma}} e^{-\mathbf{x}^T\boldsymbol\Sigma^{-1}\mathbf{x}/2}
\end{equation}
for input coordinates $x$ and covariance $\Sigma \in \mathbb{R}^{2 \times 2}$, a symmetric positive-definite matrix.

\begin{figure}[t]
\begin{center}
\adjustbox{max width=0.95\linewidth}{
\begin{tabular}{c c c}
\includegraphics[width=\linewidth]{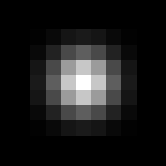} &
\includegraphics[width=\linewidth]{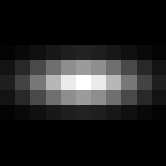} &
\includegraphics[width=\linewidth]{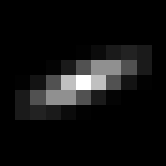} \\
\\
\resizebox{2.5in}{!}{(a) spherical} & \resizebox{2.5in}{!}{(b) diagonal} & \resizebox{1.5in}{!}{(c) full} \\
\\
\resizebox{2.5in}{!}{$\begin{bmatrix}\sigma^2 & 0 \\ 0 & \sigma^2\end{bmatrix}$ }&
\resizebox{2.5in}{!}{$\begin{bmatrix}\sigma_y^2 & 0 \\ 0 & \sigma_x^2\end{bmatrix}$} &
\resizebox{2.5in}{!}{$\begin{bmatrix}\sigma_y^2 & \rho \\ \rho & \sigma_x^2\end{bmatrix}$} \\
\end{tabular}
}
\end{center}
\vspace{-5.5mm}
\caption{
Gaussian covariances come in families with progressively richer structure:
(a) spherical covariance has one parameter for scale;
(b) diagonal covariance has two parameters for scale and aspect;
and (c) full covariance has three parameters for scale, aspect, and orientation/slant.}
\label{fig:cov-degree}
\end{figure}

The structure of the Gaussian is controlled by its covariance $\Sigma$.
Note that we are concerned with the spatial covariance, where the coordinates are considered as random variables, and not the covariance of the feature dimensions.  
Therefore the elements of the covariance matrix are $\sigma_y^2$, $\sigma_x^2$ for the y, x coordinates and $\rho$ for their correlation.
The standard, isotropic Gaussian has identity covariance $\begin{smatrix}1 & 0 \\ 0 & 1\end{smatrix}$.
There is progressively richer structure in spherical, diagonal, and full covariance: Figure \ref{fig:cov-degree} illustrates these kinds and the scale, aspect, and orientation structure they represent.

Selecting the right spatial covariance yields invariance to a given spatial transformation.
The standard Gaussian indexes scale-space, while the full covariance Gaussian indexes its \emph{affine} %
extension \cite{lindeberg1994scale-space}.
We leverage this transformation property of Gaussians to learn receptive field shape in Section \ref{sec:conv-cov} and dynamically adapt their structure for local spatially invariant filtering in Section \ref{sec:dyna-struct}.

From the Gaussian kernel $G(x, \Sigma)$ we instantiate a Gaussian filter $g_\Sigma(\cdot)$ in the standard way: (1) evaluate the kernel at the coordinates of the filter coefficients and (2) renormalize by the sum to correct for this discretization.
We decide the filter size according to the covariance by setting the half size $= \ceil{2\sigma}$ in each dimension.
This covers $\pm\;2\sigma$ to include $95\%$ of the true density no matter the covariance.
(We found that higher coverage did not improve our results.)
Our filters are always odd-sized to keep coordinates centered.

\subsection{Covariance Parameterization \& Optimization}

The covariance $\Sigma$ is symmetric positive definite, requiring proper parameterization for unconstrained optimization.
We choose the log-Cholesky parameterization \cite{pinheiro1996unconstrained} for iterative optimization because it is simple and quick to compute:
$\Sigma = U'U$ for upper-triangular $U$ with positive diagonal.
We keep the diagonal positive by storing its log, hence \emph{log}-Cholesky, and exponentiating when forming $\Sigma$.
(See \cite{pinheiro1996unconstrained} for a primer on covariance parameterization.)

Here is an example for full covariance $\Sigma$ with elements $\sigma_y^2$, $\sigma_x^2$ for the y, x coordinates and $\rho$ for their correlation:
\begin{align*}
\begin{bmatrix}\sigma_y^2 & \rho \\ \rho & \sigma_x^2\end{bmatrix}
&\gets \begin{bmatrix}+1 & -2 \\ -2 & +8\end{bmatrix}
= \begin{bmatrix}+1 & +0 \\ -2 & +2\end{bmatrix} \begin{bmatrix}+1 & -2 \\ +0 & +2\end{bmatrix} \\
\vspace{1em}
&= (\log(1), -2, \log(2)).
\end{align*}
Spherical and diagonal covariance are parameterized by fixing $\rho = 0$ and tying/untying $\sigma_y, \sigma_x$.
Note that we overload notation and use $\Sigma$ interchangeably for the covariance matrix and its log-Cholesky parameters.

Our composition learns $\Sigma$ by end-to-end optimization of structured parameters, not statistical estimation of empirical distributions.
In this way the Gaussian is determined by the task loss, and not by input statistics, as is more common.

\begin{figure}[t]
\centering
\includegraphics[width=0.95\linewidth]{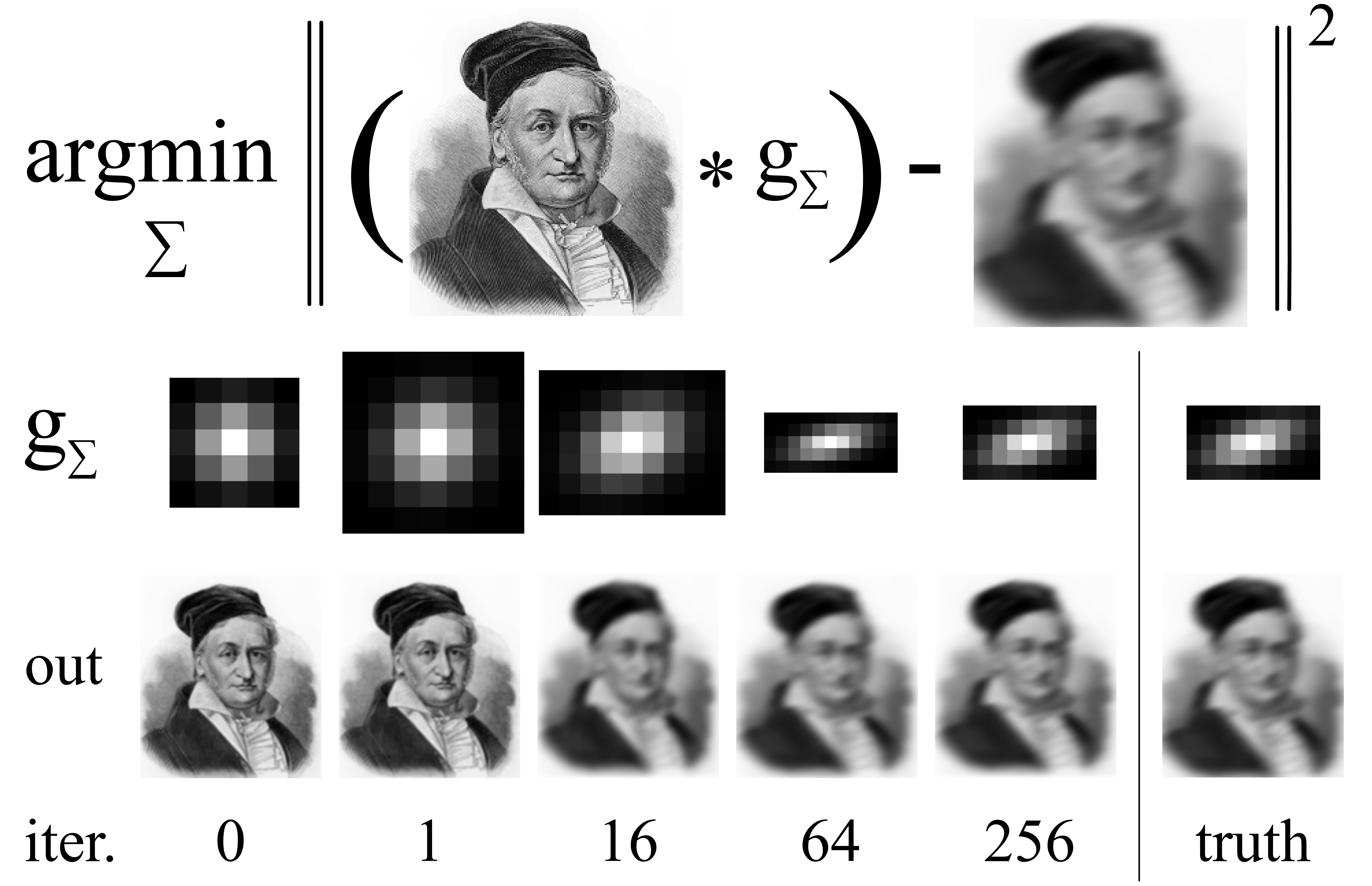}
\vspace{-3mm}
\caption{
Recovering an unknown blur by optimizing over covariance.
Gradient optimization of the structured parameters $\Sigma$ quickly converges to the true Gaussian.
Although this is a simple example, it shows the effectiveness of the Gaussian for representing scale, aspect, and orientation.
}
\label{fig:toy-blur}
\end{figure}

\subsection{Learning to Blur}

As a pedagogical example, consider the problem of optimizing covariance to reproduce an unknown blur.
That is, given a reference image and a blurred version of it, which Gaussian filter causes this blur?
Figure \ref{fig:toy-blur} shows such an optimization: from an identity-like initialization the covariance parameters quickly converge to the true Gaussian.

Given the full covariance parameterization, optimization controls scale, aspect, and orientation.
Each degree of freedom can be seen across the iterates of this example.
Had the true blur been simpler, for instance spherical, it could still be swiftly recovered in the full parameterization.

Notice how the size and shape of the filter vary over the course of optimization: this is only possible through structure.
For a Gaussian filter, its covariance is the intrinsic structure, and its coefficients follow from it.
The filter size and shape change while the dimension of the covariance itself is constant.
Lacking structure, free-form parameterization couples the number of parameters and filter size, and so cannot search over size and shape in this fashion.


\begin{figure}
\centering
\begin{subfigure}[b]{0.3\linewidth}
    \includegraphics[width=\textwidth]{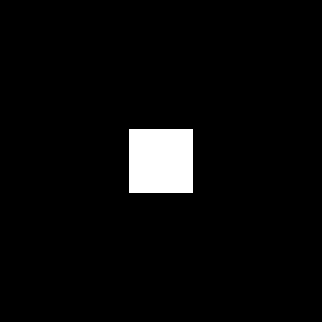}
    \caption{limit as $\sigma \to 0$}
\end{subfigure}
~
\begin{subfigure}[b]{0.3\linewidth}
    \includegraphics[width=\textwidth]{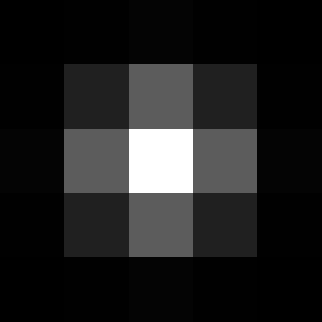}
    \caption{$\sigma < 1$}
\end{subfigure}
~
\begin{subfigure}[b]{0.3\linewidth}
    \includegraphics[width=\textwidth]{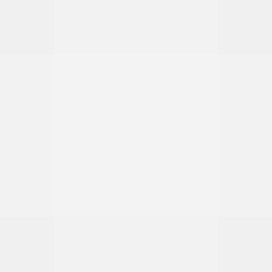}
    \caption{limit as $\sigma \to \infty$}
\end{subfigure}
\vspace{-2mm}
\caption{
Special cases of the Gaussian are helpful for differentiable model search.
(a) The identity is recovered by filtering with a delta as variance goes to zero.
(b) A smoothed delta from small variance is a good initialization to make use of pre-training.
(c) Global average pooling is recovered as variance goes to infinity.  
Each filter is normalized separately to highlight the relationship between points.
}
\label{fig:gauss-special}
\end{figure}

\section{Semi-Structured Compositional Filtering}
\label{sec:composition}

Composition and backpropagation are the twin engines of deep learning \cite{fukushima1980neocognitron,lecun1998gradient}:
composing learned linear operations with non-linearities yields deep representations.
Deep visual representations are made by composing convolutions to learn rich features and \emph{receptive fields}, which characterize the spatial extent of the features.
Although each filter might be small, and relatively simple, their composition can represent and learn large, complex receptive fields.
For instance, a stack of two $3 \times 3$ filters is effectively $5 \times 5$ but with fewer degrees of freedom ($2 \cdot 3^2$ vs. $5^2$).
Composition therefore induces factorization of the representation, and this factorization determines the generality and efficiency of the representation.

Our semi-structured composition factorizes the representation into spatial Gaussian receptive fields and free-form features.
This composition is a novel approach to making receptive field shape differentiable, low-dimensional, and decoupled from the number of parameters.
Our approach jointly learns the structured and free-form parameters while guaranteeing proper sampling for smooth signal processing.
Purely free-form filters cannot learn shape and size in this way: shape is entangled in all the parameters and size is bounded by the number of parameters.
Purely structured filters, restricted to Gaussians and their derivatives for instance, lack the generality of free-form filters.
Our factorization into structured and free-form filters is efficient for the representation, optimization, and inference of receptive fields without sacrificing the generality of features.

Receptive field size is a key design choice in the architecture of fully convolutional networks for local prediction tasks \cite{shelhamer2016fully}.
The problem of receptive field design is commonly encountered with each new architecture, dataset, or task.
Optimizing our semi-structured filters is equivalent to differentiable architecture search over receptive field size and shape.
By making this choice differentiable, we show that learning can adjust to changes in the architecture and data in Section \ref{sec:res-search}.
Trying candidate receptive fields by enumeration is expensive, whether by manual search or automated search \cite{zoph2016neural,kandasamy2018neural,liu2019darts}.
Semi-structured composition helps relieve the effort and computational burden of architecture design by relaxing the receptive field from a discrete decision into a continuous optimization.

\subsection{Composing with Convolution and Covariance}
\label{sec:conv-cov}

Our composition $f_\theta \circ g_\Sigma$ combines a free-form $f_\theta$ with a structured Gaussian $g_\Sigma$.
The computation of our composition reduces to convolution, and so it inherits the efficiency of aggressively tuned convolution implementations.
Convolution is associative, so compositional filtering of an input $I$ can be decomposed into two steps of convolution by
\begin{equation}
I * (g_\Sigma * f_\theta) = I * g_\Sigma * f_\theta.
\end{equation}
This decomposition has computational advantages.
The Gaussian step can be done by specialized filtering that harnesses separability, cascade smoothing, and other Gaussian structure.
Memory can be spared by only keeping the covariance parameters and recreating the Gaussian filters as needed (which is quick, although it is a space-time tradeoff).
Each compositional filter can always be explicitly formed by $g_\Sigma * f_\theta$ for visualization (see Figure \ref{fig:composition}) or other analysis.

Both $\theta$ and $\Sigma$ are differentiable for end-to-end learning.

How the composition is formed alters the effect of the Gaussian on the free-form filter.
Composing by convolution with the Gaussian then the free-form filter has two effects: it shapes and blurs the filter.
Composing by convolution with the Gaussian and resampling \emph{according to the covariance} purely shapes the filter.
That is, blurring and resampling first blurs with the Gaussian, and then warps the sampling points for the following filtering by the covariance.
Either operation might have a role in representation learning, so we experiment with each in Table \ref{tab:blur-vs-blur}.
In both cases the composed filter is dense, unlike a sparse filter from dilation.  

When considering covariance optimization as differentiable receptive field search, there are special cases of the Gaussian that are useful for particular purposes.
See Figure \ref{fig:gauss-special} for how the Gaussian can be reduced to the identity, initialized near the identity, or reduced to average pooling.
The Gaussian includes the identity in the limit, so our models can recover a standard networks without our composition of structure.
By initializing near the identity, we are able to augment pre-trained networks without interference, and let learning decide whether or not to make use of structure.

\begin{figure}
\begin{subfigure}[t]{0.24\linewidth}
\centering
\includegraphics[width=0.9\linewidth,page=5]{fig/dilation_blur.pdf}
\caption*{(a) input}
\end{subfigure}%
\begin{subfigure}[t]{0.24\linewidth}
\centering
\includegraphics[width=0.9\linewidth,page=4]{fig/dilation_blur.pdf}
\caption*{(b) dilated filter}
\end{subfigure}%
\begin{subfigure}[t]{0.24\linewidth}
\centering
\includegraphics[width=0.9\linewidth,page=3]{fig/dilation_blur.pdf}
\caption*{(c) output}
\end{subfigure}
\begin{subfigure}[t]{0.24\linewidth}
\centering
\includegraphics[width=0.9\linewidth,page=1]{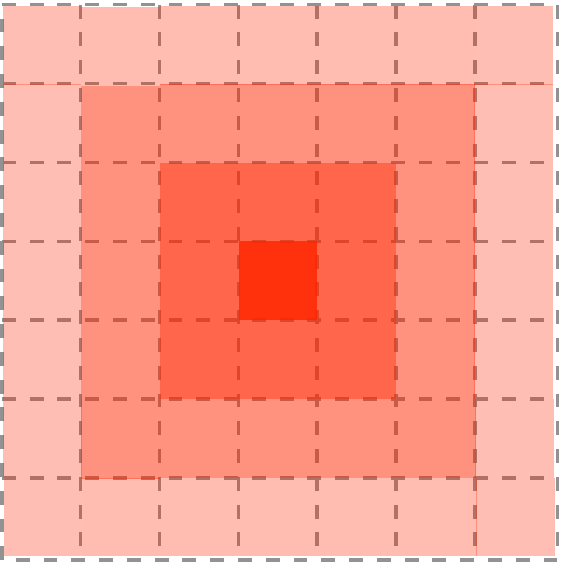}
\caption*{(d) with blur}
\end{subfigure}
\vspace{-2mm}
\caption{
Blurring prevents aliasing or ``gridding'' artifacts by smoothing dilation to respect the sampling theorem.
}
\label{fig:dilation}
\end{figure}

\minisection{Blurring for Smooth Signal Processing}
Blurring (and resampling) by the covariance guarantees proper sampling for correct signal processing.
It synchronizes the degree of smoothing and the sampling rate to avoid aliasing.
Their combination can be interpreted as a smooth, circular extension of dilated convolution \cite{chen2015semantic,yu2015multi} or as a smooth, affine restriction of deformable convolution \cite{dai2017deformable}.
Figure \ref{fig:dilation} contrasts dilation with blurring \& resampling.
For a further perspective, note this combination is equivalent to downsampling/upsampling with a Gaussian before/after convolving with the free-form filter.

Even without learning the covariance, blurring can improve dilated architectures.
Dilation is prone to gridding artifacts \cite{yu2017dilated,wang2018understanding}.
We identify these artifacts as aliasing caused by the spatial sparsity of dilated filters.
We fix this by smoothing with standard deviation proportional to the dilation rate.
Smoothing when subsampling is a fundamental technique in signal processing to avoid aliasing \cite{oppenheim2009discrete-time}, and the combination serves as a simple alternative to the careful re-engineering of dilated architectures.
Improvements from blurring dilation are reported in Table \ref{tab:dilation}.

\minisection{Compound Gaussian Structure}
Gaussian filters have a special compositional structure we can exploit: cascade smoothing.
Composing a Gaussian $g_\Sigma$ with a Gaussian $g_{\Sigma'}$ is still Gaussian with covariance $\Sigma + \Sigma'$.
This lets us efficiently assemble \emph{compound} receptive fields made of multiple Gaussians.
Center-surround \cite{kuffler1953discharge} receptive fields, which boost contrast, can be realized by such a combination as Difference-of-Gaussian \cite{rodieck1965analysis} (DoG) filters, which subtract a larger Gaussian from a smaller Gaussian.
Our joint learning of their covariances tunes the contrastive context of the receptive field, extending \cite{ding2018context} which learns contrastive filters with fixed receptive field sizes.

\minisection{Design Choices}
Having defined our semi-structured composition, we cover the design choices involved in its application.
As a convolutional composition, it can augment any convolution layer in the architecture.
We focus on including our composition in late, deep layers to show the effect without much further processing.
We add compositional filtering to the output and decoder layers of fully convolutional networks because the local tasks they address rely on the choice of receptive fields.  

Having decided where to compose, we must decide how much structure to compose.
There are degrees of structure, from minimal structure, where each layer or stage has only one shared Gaussian, to dynamic structure, where each receptive field has its own structure that varies with the input.
In between there is channel structure, where each free-form filter has its own Gaussian shared across space, or multiple structure, where each layer or filter has multiple Gaussians to cover different shapes.
We explore minimal structure and dynamic structure in order to examine the effect of composition for static and dynamic inference, and leave the other degrees of structure to future work.

\begin{figure}[t]
  \begin{subfigure}[t]{0.33\linewidth}
   \centering
   \includegraphics[width=0.9\linewidth,page=9]{fig/deform_gauss.pdf}
   \caption*{(a)\\$3 \times 3$ filter}
  \end{subfigure}%
  \begin{subfigure}[t]{0.33\linewidth}
   \centering
   \includegraphics[width=0.9\linewidth,page=8]{fig/deform_gauss.pdf}
   \caption*{(b)\\deformable \cite{dai2017deformable}}
  \end{subfigure}%
  \begin{subfigure}[t]{0.33\linewidth}
   \centering
   \includegraphics[width=0.9\linewidth,page=6]{fig/deform_gauss.pdf}
   \caption*{(c)\\standard Gauss.}
  \end{subfigure}

  \begin{subfigure}[t]{0.33\linewidth}
   \centering
   \includegraphics[width=0.9\linewidth,page=5]{fig/deform_gauss.pdf}
   \caption*{(d)\\spherical Gauss.}
  \end{subfigure}%
  \begin{subfigure}[t]{0.33\linewidth}
   \centering
   \includegraphics[width=0.9\linewidth,page=3]{fig/deform_gauss.pdf}
   \caption*{(e)\\diagonal Gauss.}
  \end{subfigure}%
  \begin{subfigure}[t]{0.33\linewidth}
   \centering
   \includegraphics[width=0.9\linewidth,page=1]{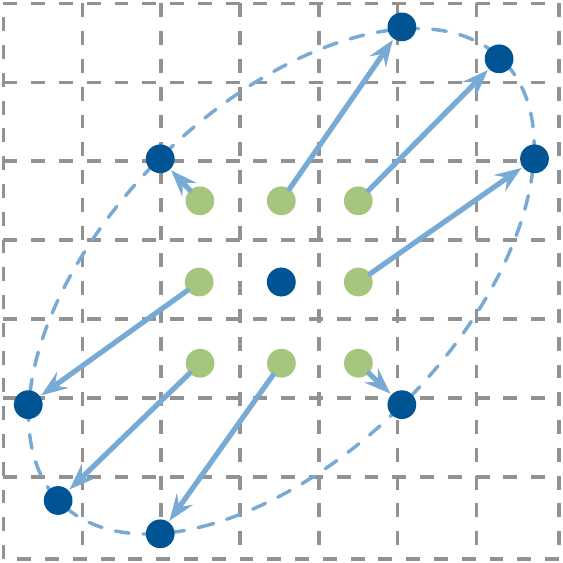}
   \caption*{(f)\\full Gauss.}
  \end{subfigure}

\vspace{-2mm}
\caption{
Gaussian deformation (c-f) structures dynamic receptive fields by controlling the sampling points (blue) through the covariance.
The low-dimensionality of covariance is more efficient than free-form deformation (b) for learning and inference.
Although it is less general, it still expresses a variety of shapes.
}
\label{fig:gauss-deform}
\end{figure}

\subsection{Dynamic Gaussian Structure}
\label{sec:dyna-struct}

Semi-structured composition learns a rich family of receptive fields, but visual structure is richer still, because structure locally varies while our filters are fixed.
Even a single image contains variations in scale and orientation, so one-size-and-shape-fits-all structure is suboptimal.
\emph{Dynamic} inference replaces static, global parameters with dynamic, local parameters that are inferred from the input to adapt to these variations.
Composing with structure by convolution cannot locally adapt, since the filters are constant across the image.
We can nevertheless extend our composition to dynamic structure by representing local covariances and instantiating local Gaussians accordingly.
Our composition makes dynamic inference efficient by decoupling low-dimensional, Gaussian structure from high-dimensional, free-form filters.

There are two routes to dynamic Gaussian structure: local filtering and deformable sampling.
Local filtering has a different filter kernel for each position, as done by dynamic filter networks \cite{de-brabandere2016dynamic}.
This ensures exact filtering for dynamic Gaussians, but is too computationally demanding for large-scale recognition networks.
Deformable sampling adjusts the position of filter taps by arbitrary offsets, as done by deformable convolution \cite{dai2017deformable}.
We exploit deformable sampling to dynamically form sparse approximations of Gaussians.

We constrain deformable sampling to Gaussian structure by setting the sampling points through covariance.
Figure \ref{fig:gauss-deform} illustrates these Gaussian deformations.
We relate the default deformation to the standard Gaussian by placing one point at the origin and circling it with a ring of eight points on the unit circle at equal distances and angles.
We consider the same progression of spherical, diagonal, and full covariance for dynamic structure.
This low-dimensional structure differs from the high degrees of freedom in a dynamic filter network, which sets free-form filter parameters, and deformable convolution, which sets free-form offsets.
In this way our semi-structured composition requires only a small, constant number of covariance parameters independent of the sampling resolution and the kernel size $k$, while deformable convolution has constant resolution and requires $2k^2$ offset parameters for a $k \times k$ filter.

To infer the local covariances, we follow the deformable approach \cite{dai2017deformable}, and learn a convolutional regressor for each dynamic filtering step.
The regressor, which is simply a convolution layer, first infers the covariances which then determine the dynamic filtering that follows.
The low-dimensional structure of our dynamic parameters makes this regression more efficient than free-form deformation, as it only has three outputs for each full covariance, or even just one for each spherical covariance.
Since the covariance is differentiable, the regression is learned end-to-end from the task loss without further supervision.

We experiment with dynamic structure in Section \ref{sec:res-dyna}.

\section{Experiments}
\label{sec:res}

We experiment with the local task of semantic segmentation, because our method learns the size and shape of local receptive fields.
As a recognition task, semantic segmentation requires a balance between local scope, to infer where, and global scope, to infer what.
Existing approaches must take care with receptive field design, and their experimental development takes significant model search.

\minisection{Data}
CityScapes \cite{cordts2016cityscapes} is a challenging dataset of varied urban scenes from the perspective of a car-mounted camera.
We follow the standard training and evaluation protocols and train/validation splits, with $2,975$ finely-annotated training images and $500$ validation images.
We score results by the common intersection-over-union metric---the intersection of predicted and true pixels divided by their union then averaged over classes---on the validation set.
We evaluate the network itself without post-processing, test-time augmentation, or other accessories to isolate the effect of receptive field learning.

\minisection{Architecture and Optimization}
For backbones we choose strong fully convolutional networks derived from residual networks \cite{he2016deep}.
The dilated residual net (DRN) \cite{yu2017dilated} has high resolution and receptive field size through dilation.
Deep layer aggregation (DLA)  \cite{yu2018deep} fuses layers by hierarchical and iterative skip connections.
We also define a ResNet-34 backbone as a simple architecture of the kind used for ablations and exploratory experiments.
Together they are representative of common architectural patterns in state-of-the-art fully convolutional networks.

We train our models by stochastic gradient descent for $240$ epochs with momentum $0.9$, batch size 16, and weight decay $10^{-4}$.
Training follows the ``poly'' learning rate schedule \cite{chen2018deeplab, zhao2017pyramid} with initial rate $0.01$.
The input images are cropped to $800\times800$ and augmented by random scaling within $[0.5, 2]$, random rotation within 10 degrees, and random color distortions as in ~\cite{howard2013some}.
We train with synchronized, in-place batch normalization \cite{rota2018place}.
For fair comparison, we reproduce the DRN and DLA baselines in our same setting, which improves on their reported results.

\minisection{Baselines}
The chosen DRN and DLA architectures are strong methods on their own, but they can be further equipped for learning global spatial transformations and local deformations.
Spatial transformer networks \cite{jaderberg2015spatial} and deformable convolution \cite{dai2017deformable} learn dynamic global/local transformations respectively.
Spatial transformers serve as a baseline for structure, because they are restricted to a parametric class of transformations.
Deformable convolution serves as a baseline for local, dynamic inference without structure.
For comparison in the static setting, we simplify both methods to instead learn static transformations.

Naturally, because our composition is carried out by convolution (for static inference), we compare to the baseline of including a free-form convolution layer on its own.

We will release code and reference models for our static and dynamic compositional filtering methods.

\subsection{Learning Semi-Structured Filters}
\label{sec:res-filters}

We first show that semi-structured compositional filtering improves the accuracy of strong fully convolutional networks.
We then examine how to best implement our composition and confirm the value of smooth signal processing.

\minisection{Augmenting Backbone Architectures}
Semi-structured filtering improves the accuracy of strong fully convolutional networks.
We augment the last, output stage with a single instance of our composition and optimize end-to-end.
See Table \ref{tab:sota} for the accuracies of the backbones, baselines, and our filtering.
Static composition by convolution improves on the backbone by 1-2 points, and dynamic composition boosts the improvement to 4 points (see Section \ref{sec:res-dyna}).

Our simple composition improves on the accuracy of the static receptive fields learned by a spatial transformer and deformable convolution.
Spatial transformers and our static composition each learn a global transformation, but our Gaussian parameterization is more effectively optimized.
Deformable convolution learns local receptive fields, but its free-form parameterization takes more computation and memory.
Our edition of DoG, which learns the surround size, improves the accuracy a further $0.5$ point.

Note that the backbones are agressively-tuned architectures which required significant model search and engineering effort.
Our composition is still able to deliver improvement through learning without further engineering.
In the next subsection, we show that joint optimization of our composition does effective model search when the chosen architecture is suboptimal.

\begin{table}
\begin{center}
\begin{tabular}{l|ccc}
\hline
method & IU \\
\hline\hline
DRN-A \cite{yu2017dilated}            & 72.4 \\
+ $3\times3$ Conv.                             & 72.9 \\
+ STN (static) \cite{jaderberg2015spatial}     & 70.5 \\
+ Deformable (static) \cite{dai2017deformable} & 72.2 \\
+ Composition (ours)                           & \bf{73.5} \\
\hline
+ CCL \cite{ding2018context}                   & 73.1 \\
+ DoG (ours)                                   & \bf{74.1} \\
\hline\hline
DLA-34 \cite{yu2018deep}                       & 76.1 \\
+ Composition (ours)                           & \bf{78.2} \\
\hline
\end{tabular}
\end{center}
\caption{
Our composition improves the accuracy of strong, carefully designed architectures.
}
\label{tab:sota}
\end{table}

\minisection{How to Compose}
As explained in Section \ref{sec:conv-cov}, we can compose with a Gaussian structured filter by blurring alone or blurring and resampling.
As either can be learned end-to-end, we experiment with both and report their accuracies in Table \ref{tab:blur-vs-blur}.
From this comparison we choose blurring and resampling for the remainder of our experiments.

\begin{table}
\begin{center}
\begin{tabular}{l|c}
\hline
method & IU \\
\hline\hline
ResNet-34             & 64.8 \\
+ Blur                & 66.3 \\
+ Blur-Resample       & \bf{68.1} \\
\hline
+ DoG Blur            & 70.3 \\
+ DoG Blur-Resample   & \bf{71.4} \\
\hline\hline
DRN-A \cite{yu2017dilated} & 72.4 \\
+ Blur                     & 72.2 \\
+ Blur-Resample            & \bf{73.5} \\
\hline
\end{tabular}
\end{center}
\caption{
For semi-structured composition, blurring + resampling improves on blurring alone.
This holds for the simple composition of a Gaussian and free-form filter, and the compound composition of Difference of Gaussian filtering.
}
\label{tab:blur-vs-blur}
\end{table}

\begin{table}
\begin{center}
\begin{tabular}{l|c}
\hline
method & IU \\
\hline\hline
DRN-A \cite{yu2017dilated}     & 72.4 \\
\hline
w/ CCL \cite{ding2018context}  & 73.1 \\
+ Blur                         & \bf{74.0} \\
\hline
w/ ASPP \cite{chen2018deeplab} & 74.1 \\
+ Blur                         & \bf{74.3} \\
\hline
\end{tabular}
\end{center}
\caption{
Blurred dilation respects the sampling theorem by smoothing in proportion to the dilation rate.
Blurring in this way gives a small boost in accuracy.
}
\label{tab:dilation}
\end{table}

\minisection{Blurred Dilation}
To isolate the effect of blurring without learning, we smooth dilation with a blur proportional to the dilation rate.
CCL \cite{ding2018context} and ASPP \cite{chen2018deeplab} are carefully designed dilation architectures for context modeling, but neither blurs before dilating.
Improvements from blurred dilation are reported in Table \ref{tab:dilation}.
Although the gains are small, this establishes that smoothing can help.
This effect should only increase with dilation rate.

The small marginal effect of blurring without learning shows that most of our improvement is from joint optimization of our composition and dynamic inference.


\subsection{Differentiable Receptive Field Search}
\label{sec:res-search}

Our composition makes local receptive fields differentiable in a low-dimensional, structured parameterization.
This turns choosing receptive fields into a task for learning, instead of designing or manual searching.
We demonstrate that this differentiable receptive field search is able to adjust for changes in the architecture and data.
Table \ref{tab:search} shows how receptive field optimization counteracts the reduction of the architectural receptive field size and the enlargement of the input.
These controlled experiments, while simple, reflect a realistic lack of knowledge in practice: for a new architecture or dataset, the right design is unknown.

For these experiments we include our composition in the last stage of the network and \emph{only optimize this stage}.
We do this to limit the scope of learning to the joint optimization of our composition, since then any effect is only attributable to the composition itself.
We verify that end-to-end learning further improves results, but controlling for it in this way eliminates the possibility of confounding effects.

In the extreme, we can do \emph{structural} fine-tuning by including our composition in a pre-trained network and only optimizing the covariance.
When fine-tuning the structure alone, optimization either reduces the Gaussian to a delta, doing no harm, or slightly enlarges the receptive field, giving a one point boost.
Therefore the special case of the identity, as explained in Figure \ref{fig:gauss-special}, is learnable in practice.
This shows that our composition helps or does no harm, and further supports the importance of jointly learning the composition as we do.

\begin{table}
\begin{center}
\begin{tabular}{l|cccc}
\hline
method & no. params & epoch & IU & $\Delta$ \\
\hline
\hline
DRN-A \cite{yu2017dilated}          & many & 240 & 72.4 & 0 \\
\hline
\hline
\multicolumn{5}{l}{Smaller Receptive Field} \\
\hline\hline
ResNet-34                           & many & 240 & 64.8 & -7.6 \\
+ $3\times3$ Conv.                  & some & +20 & 65.8 & -6.6 \\
+ Composition                       & some & +20  & 68.1 & -4.6 \\
+ DoG                               & some & +20  & 68.9 & -3.5 \\
\dots End-to-End                    & many & 240  & \bf{71.4} & -0.8 \\
\hline
\hline
\multicolumn{5}{l}{and $2\times$ Enlarged Input} \\
\hline\hline
ResNet-34                          & many & 240  & 56.2 & -16.2 \\
+ $3\times3$ Conv.                 & some & +20  & 56.7 & -15.7 \\
+ Composition                      & some & +20  &  57.8 & -14.6 \\
+ DoG                              & some & +20  & 62.7 & -9.7 \\
\dots End-to-End                   & many & 240  & \bf{66.5} & -5.9 \\
\hline
\end{tabular}
\end{center}
\caption{
  Adjusting to architecture and data by differentiable receptive field search.
  When the architectural receptive field is reduced, the learned covariance compensates to enlarge it.
  When the input is additionally enlarged $2\times$, the learned covariance grows further still.
}
\label{tab:search}
\end{table}

\subsection{Dynamic Inference of Gaussian Structure}
\label{sec:res-dyna}

Learning the covariance optimizes receptive field size and shape.
Dynamic inference of the covariance takes this a step further, and adaptively adjusts receptive fields to vary with the input.
By locally regressing the covariance, our approach can better cope with factors of variation \emph{within} an image, and do so efficiently through structure.

\begin{table}
\begin{center}
\begin{tabular}{l|ccc}
\hline
\multicolumn{4}{l}{Cityscapes Validation} \\
\hline
method & dyn.? & \pbox{5em}{no. dyn.\\params} & IU \\
\hline\hline
DRN-A \cite{yu2017dilated}                        &            & -      & 72.4 \\
+ Static Composition (ours)                       &            & -      & 73.5 \\
+ Gauss. Deformation (ours)                       & \checkmark & 1      & \bf{76.6} \\
+ Free-form Deformation \cite{dai2017deformable}  & \checkmark & $2k^2$ & \bf{76.6} \\
\hline
ResNet-34                                         &            & -      & 64.8 \\
+ Static Composition (ours)                       &            & -      & 68.1 \\
+ Gauss. Deformation (ours)                       & \checkmark & 1      & 74.2 \\
+ Free-form Deformation \cite{dai2017deformable}  & \checkmark & $2k^2$ & \bf{75.1} \\
\hline\hline
\multicolumn{4}{l}{Cityscapes Test} \\
\hline
\hline
DRN-A \cite{yu2017dilated}                        &            & -      & 71.2 \\
+ Gauss. Deformation (ours)                       & \checkmark & 1      & \bf{74.3} \\
+ Free-form Deformation \cite{dai2017deformable}  & \checkmark & $2k^2$ & 73.6 \\
\hline
\end{tabular}
\end{center}
\caption{
Dynamic Gaussian deformation reduces parameters, improves computational efficiency, and rivals the accuracy of free-form deformation.
Even restricting the deformation to scale by spherical covariance suffices to nearly equal the free-form accuracy.
}
\label{tab:dyna-struct}
\end{table}

We compare our Gaussian deformation with free-form deformation in Table \ref{tab:dyna-struct}.
Controlling deformable convolution by Gaussian structure improves efficiency while preserving accuracy to within one point.
While free-form deformations are more general in principle, in practice there is a penalty in efficiency.
Recall that the size of our structured parameterization is independent of the free-form filter size.
On the other hand unstructured deformable convolution requires $2k^2$ parameters for a $k \times k$ filter.

\begin{figure*}[ht]
\begin{center}
\adjustbox{max width=\linewidth}{
\begin{tabular}{c c c c}
\includegraphics[width=\linewidth]{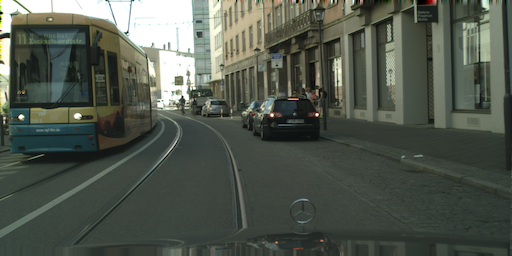} &
\includegraphics[width=\linewidth]{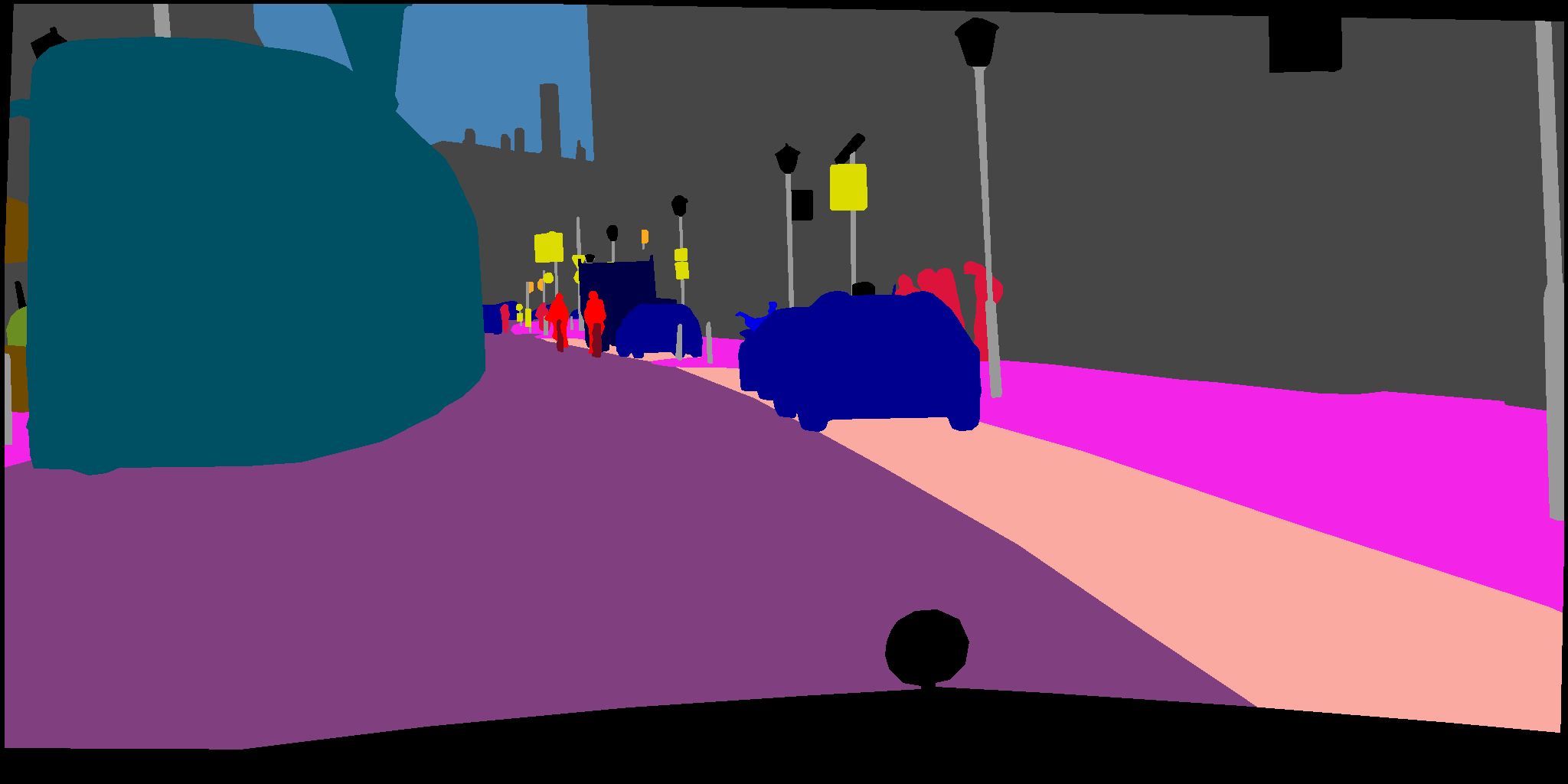} &
\includegraphics[width=\linewidth]{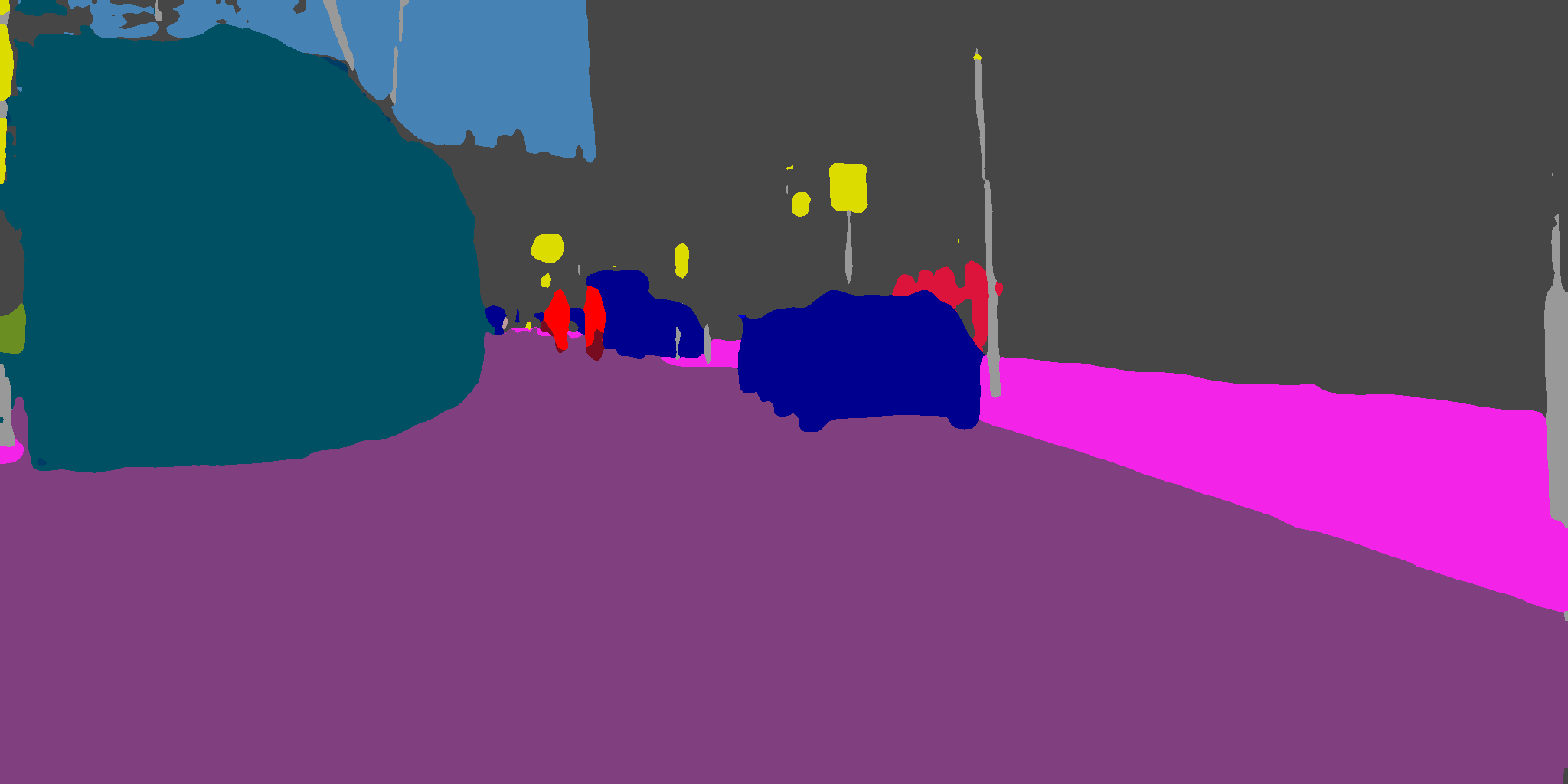} &
\includegraphics[width=\linewidth]{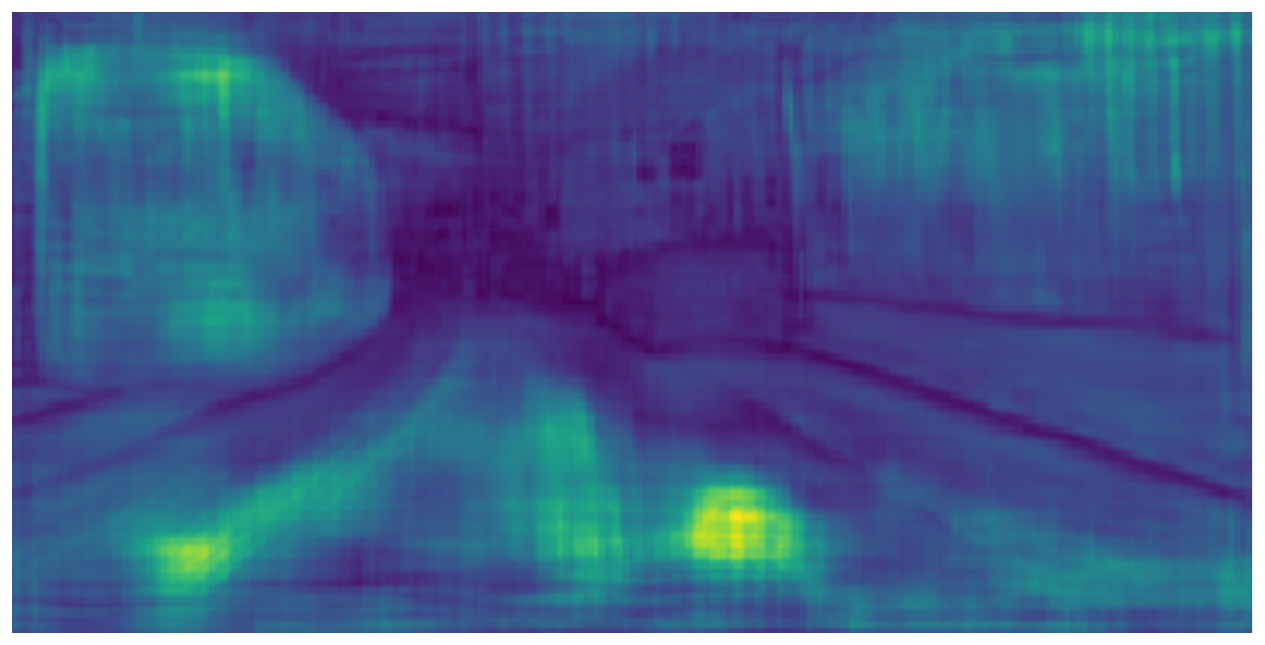} \\
\includegraphics[width=\linewidth]{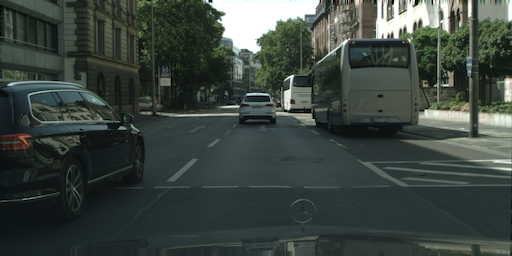} &
\includegraphics[width=\linewidth]{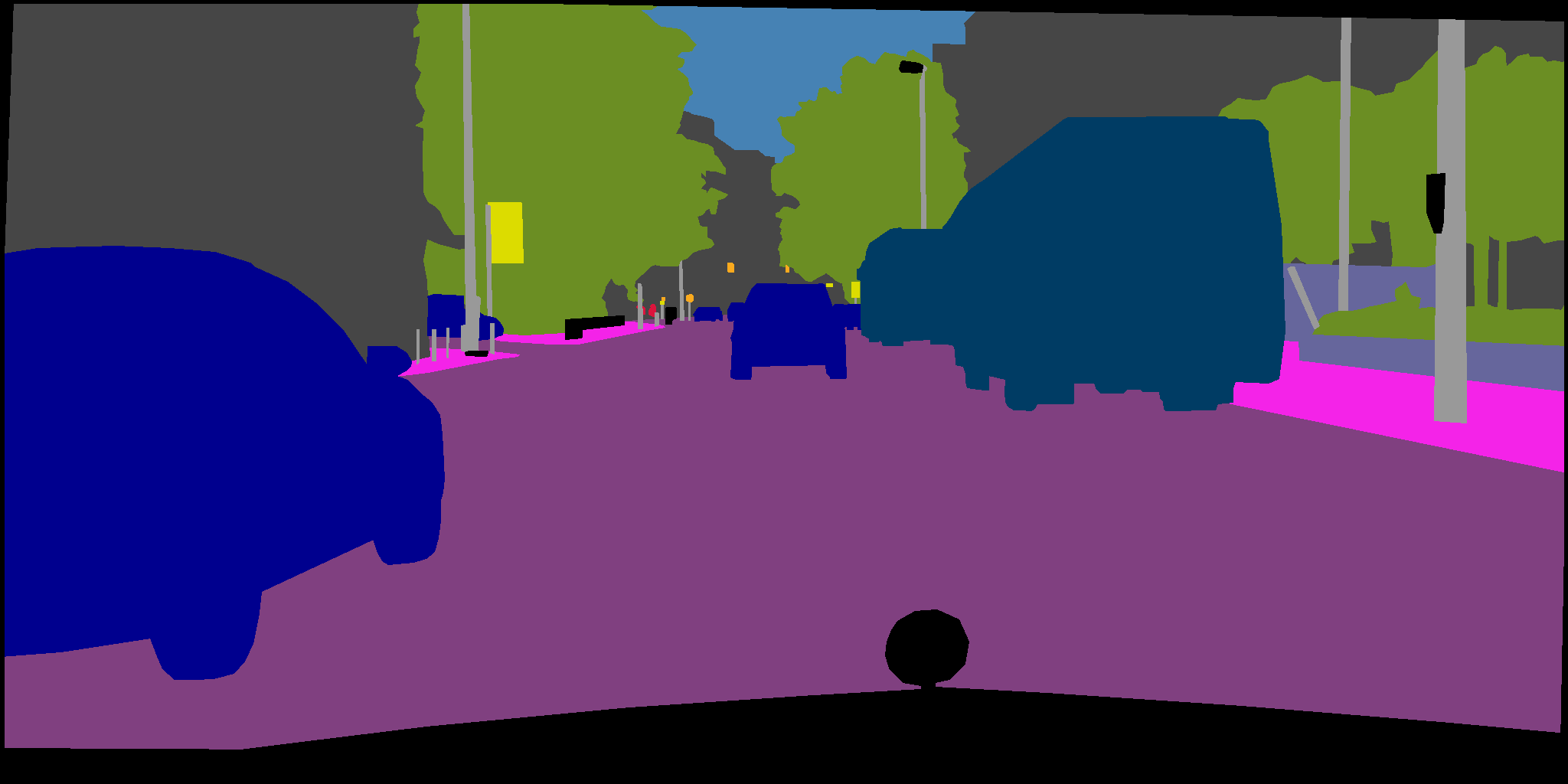} &
\includegraphics[width=\linewidth]{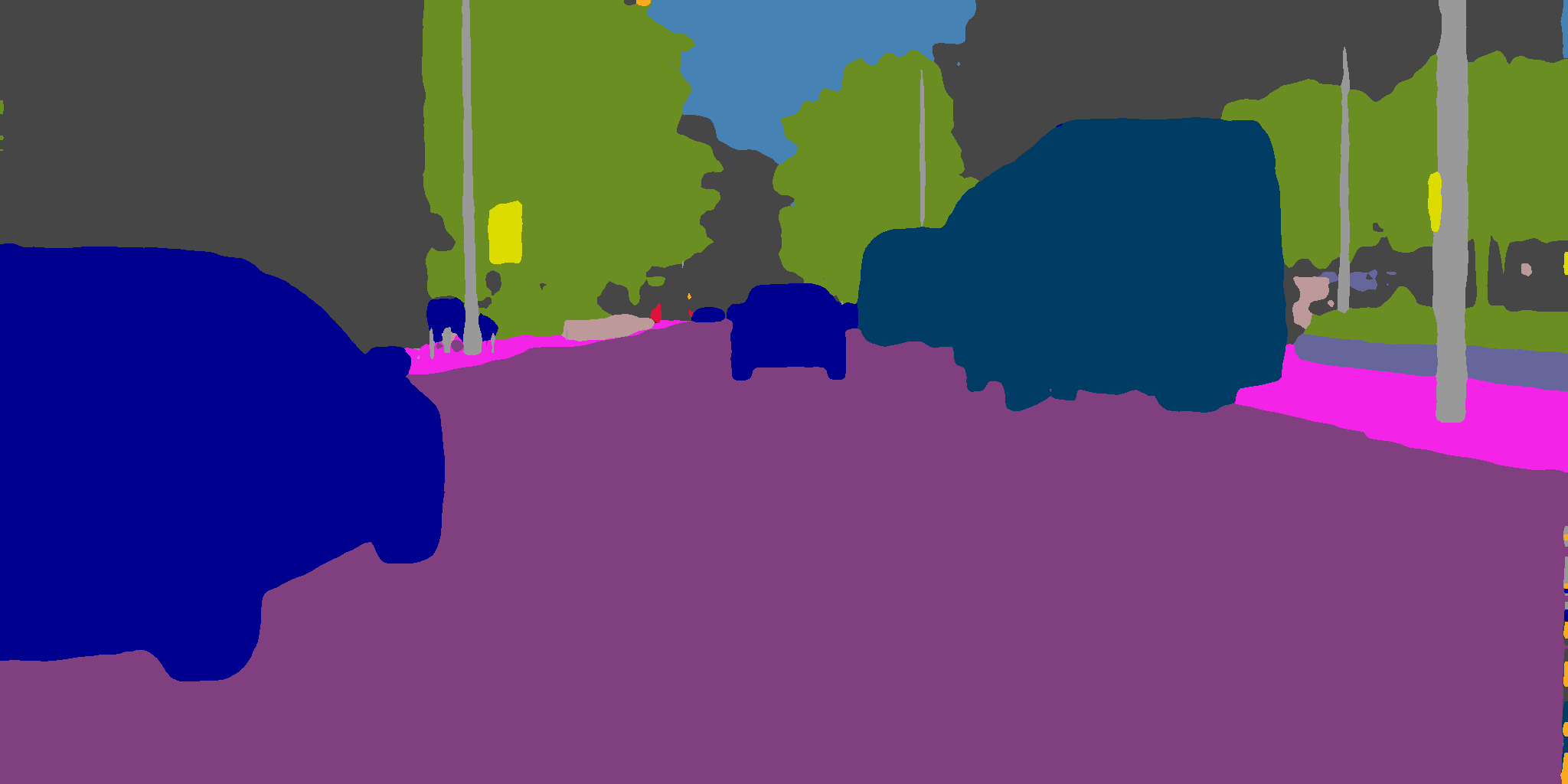} &
\includegraphics[width=\linewidth]{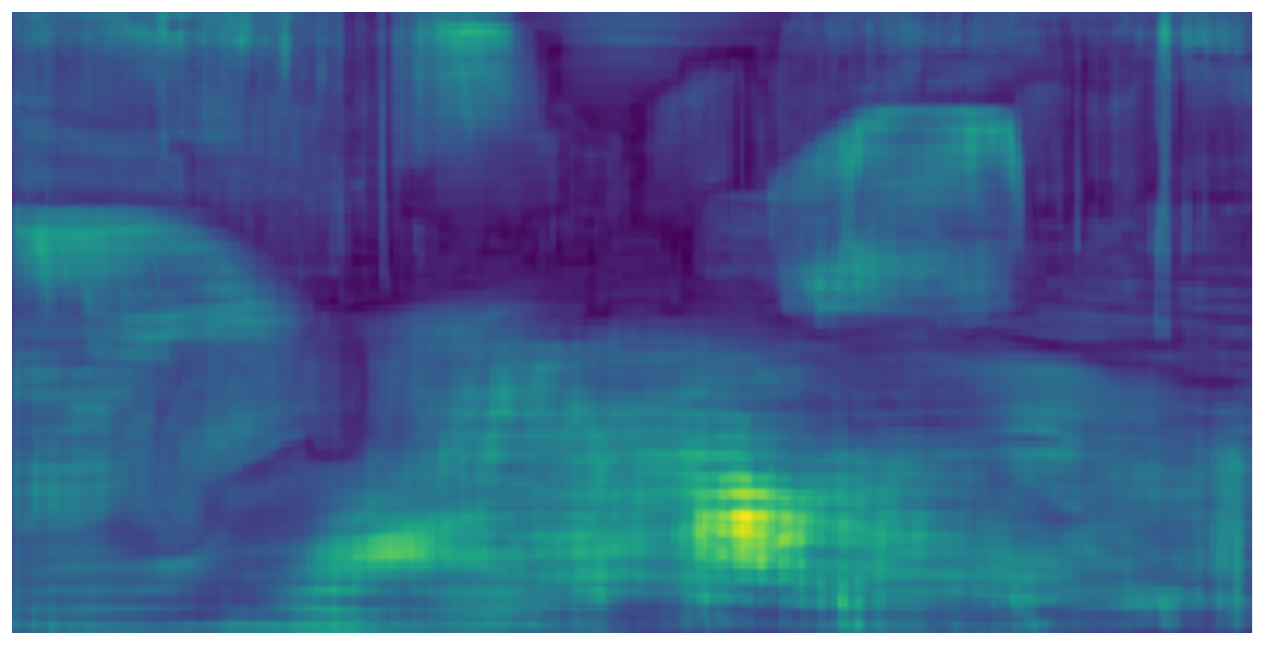} \\
\includegraphics[width=\linewidth]{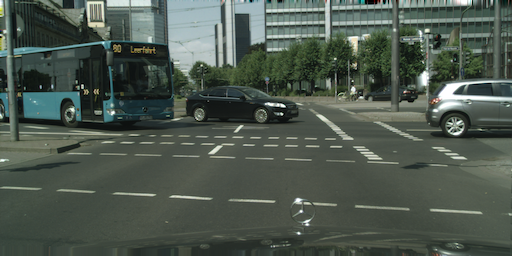} &
\includegraphics[width=\linewidth]{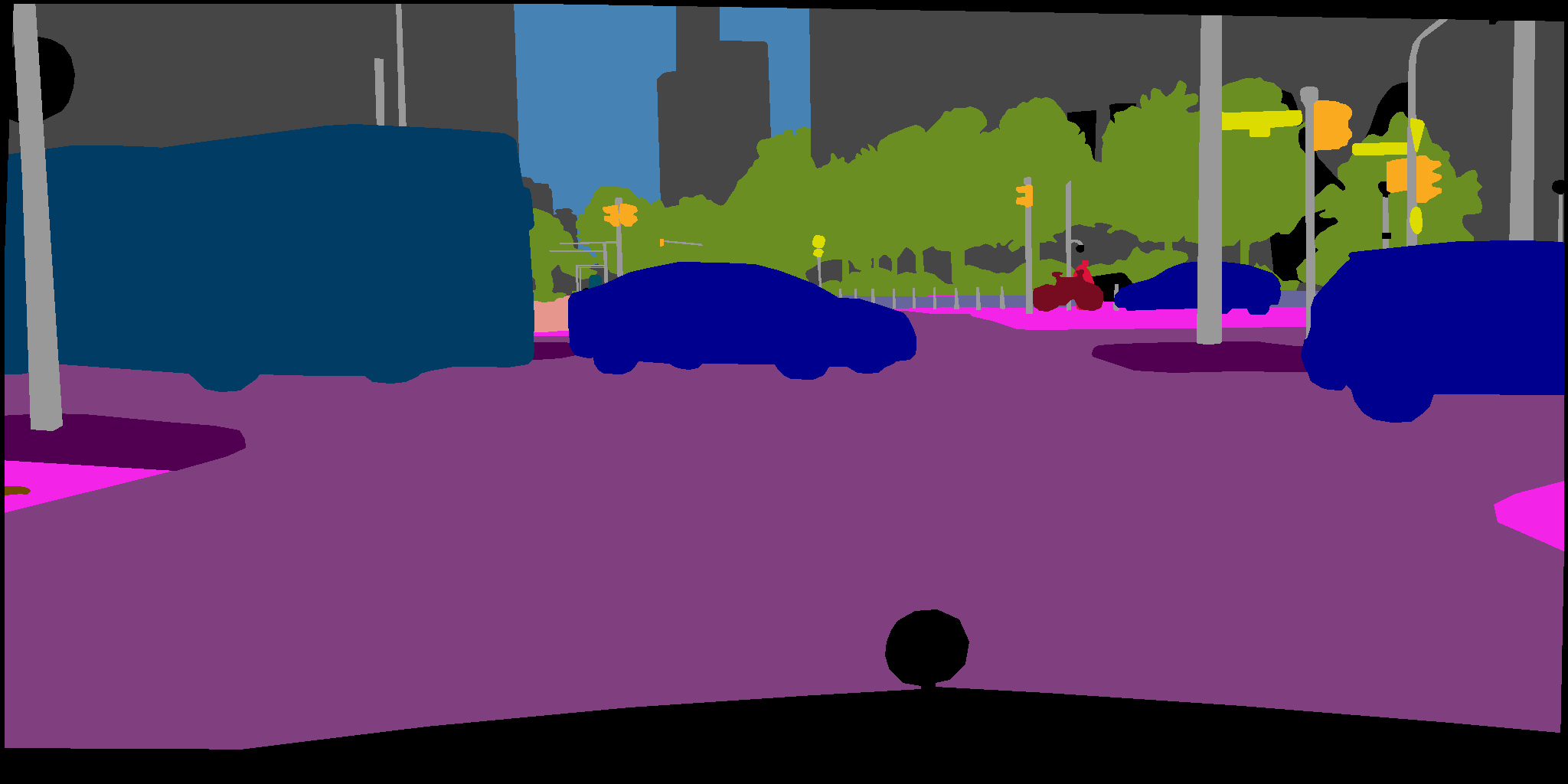} &
\includegraphics[width=\linewidth]{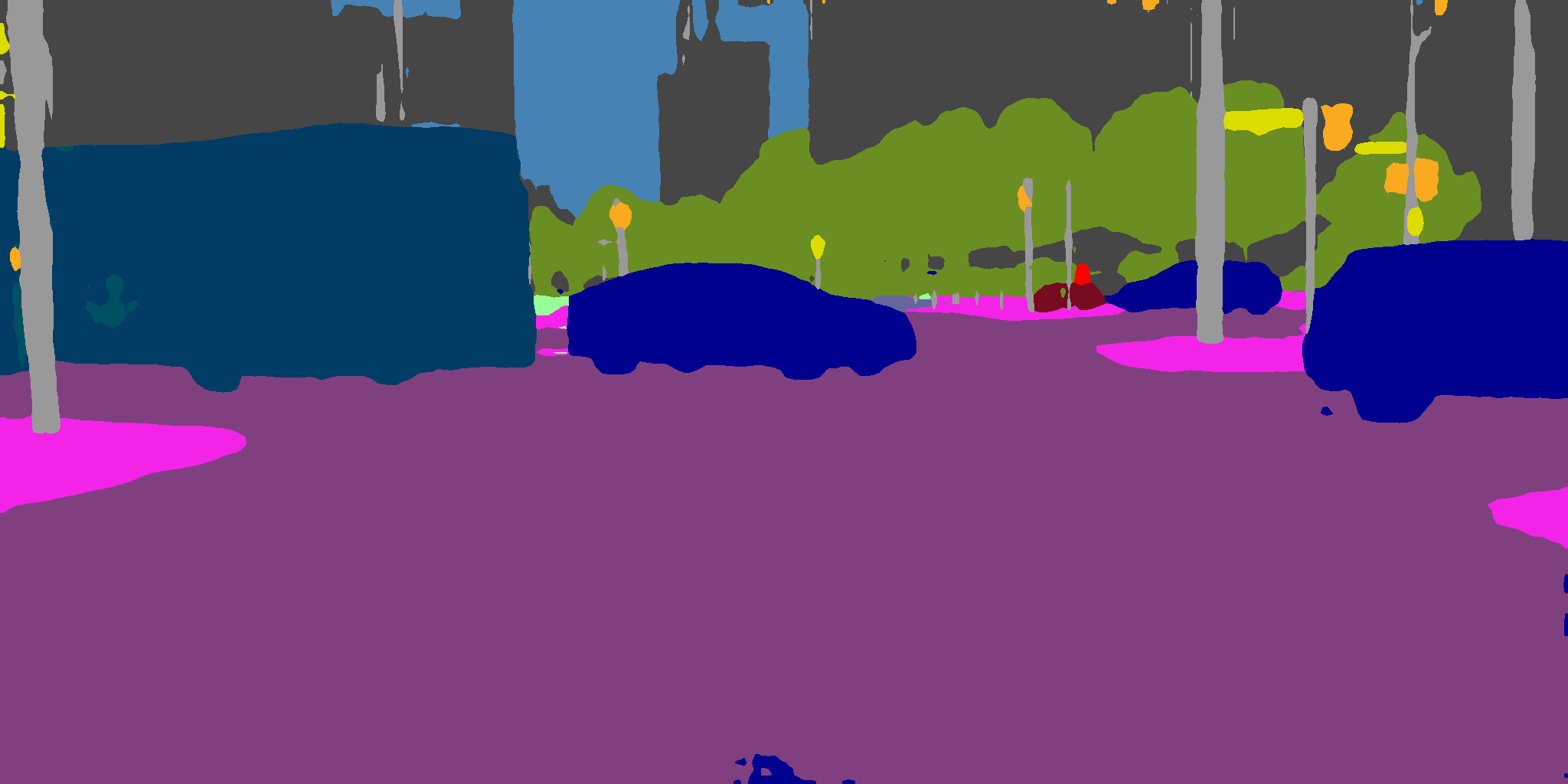} &
\includegraphics[width=\linewidth]{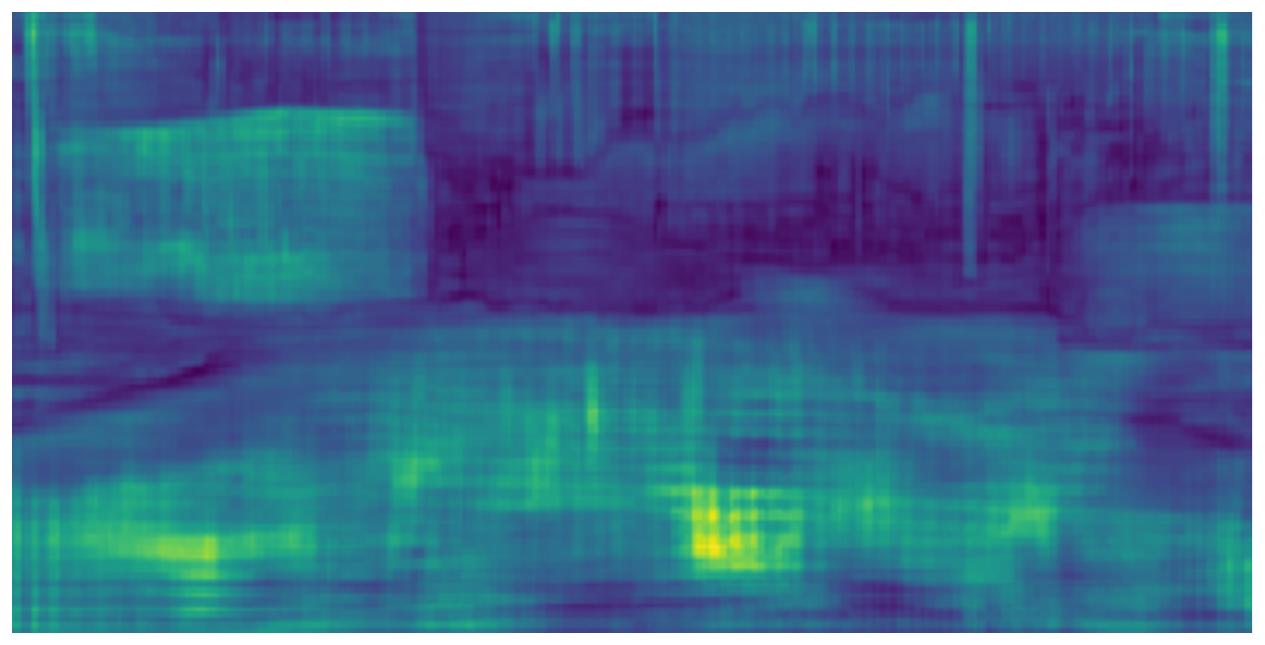} \\
\includegraphics[width=\linewidth]{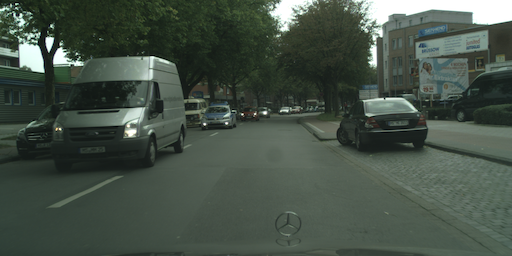} &
\includegraphics[width=\linewidth]{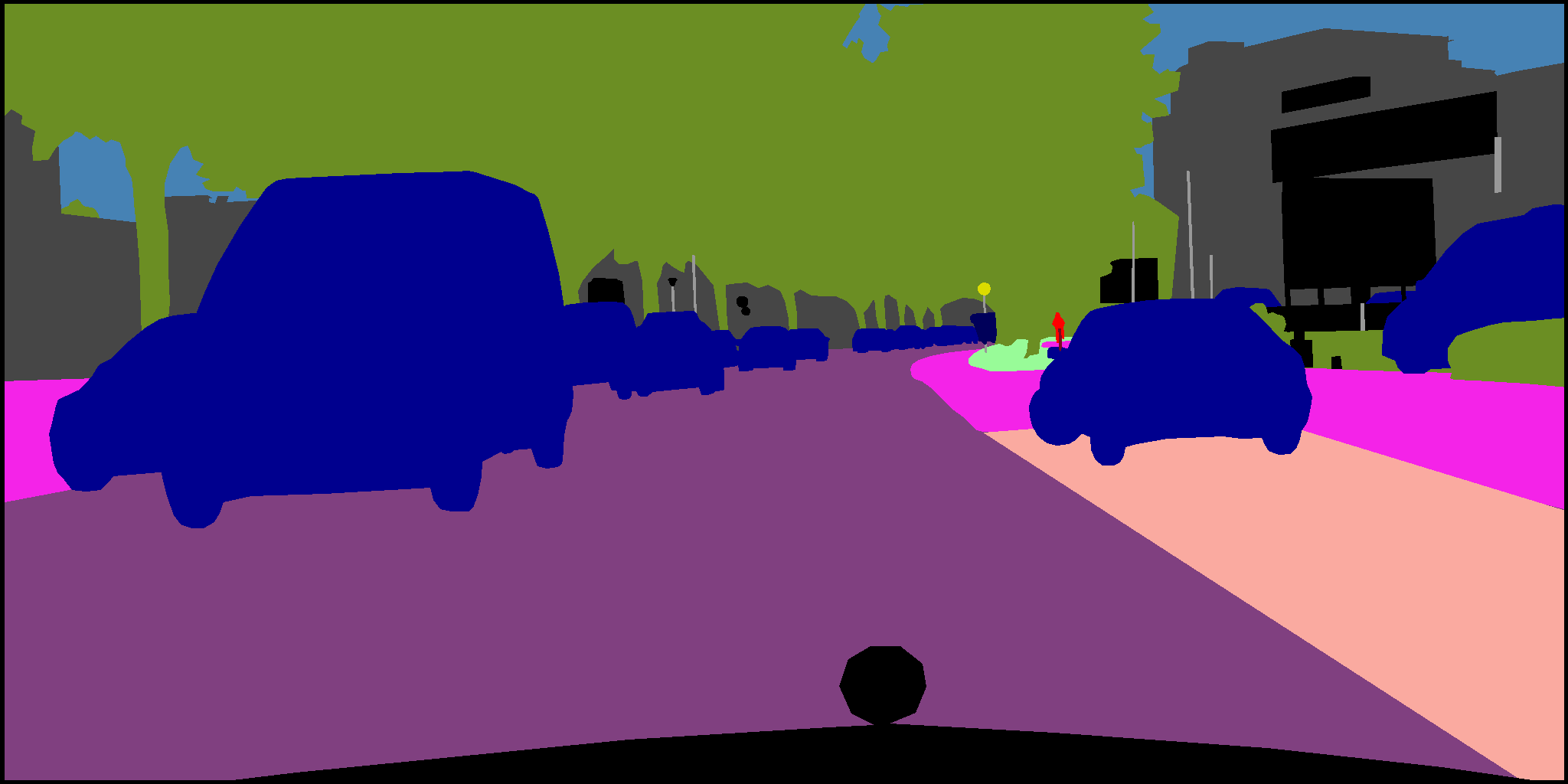} &
\includegraphics[width=\linewidth]{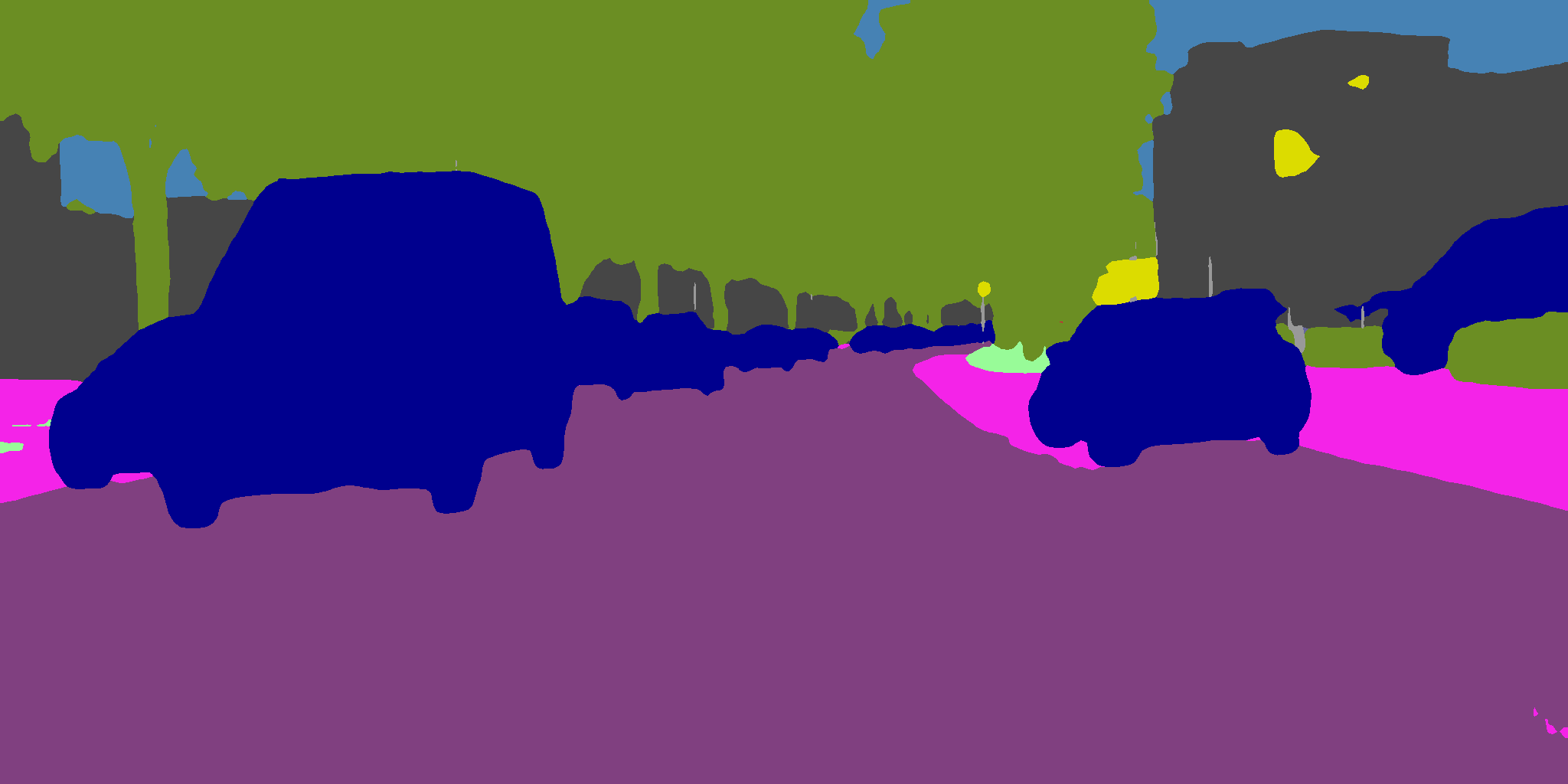} &
\includegraphics[width=\linewidth]{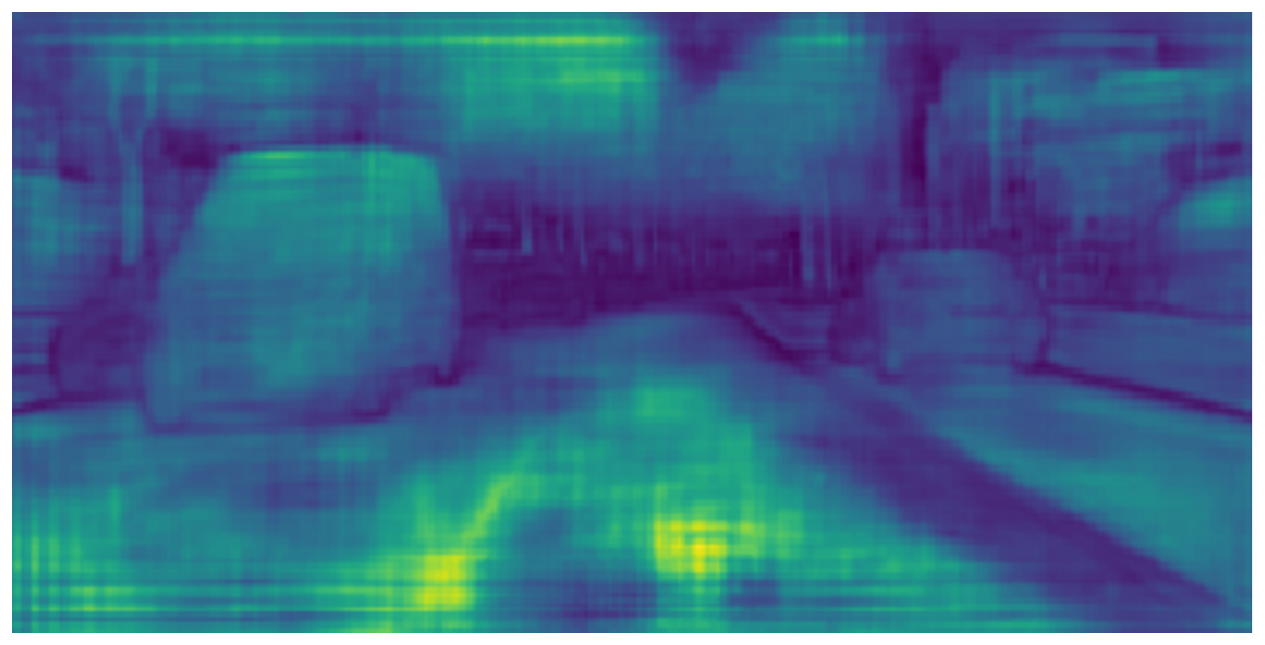} \\
\includegraphics[width=\linewidth]{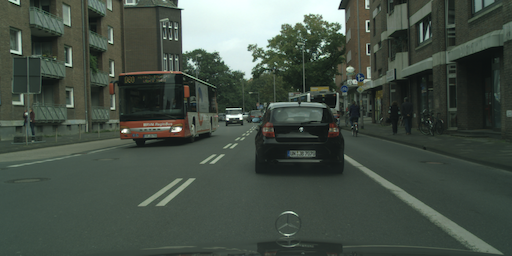} &
\includegraphics[width=\linewidth]{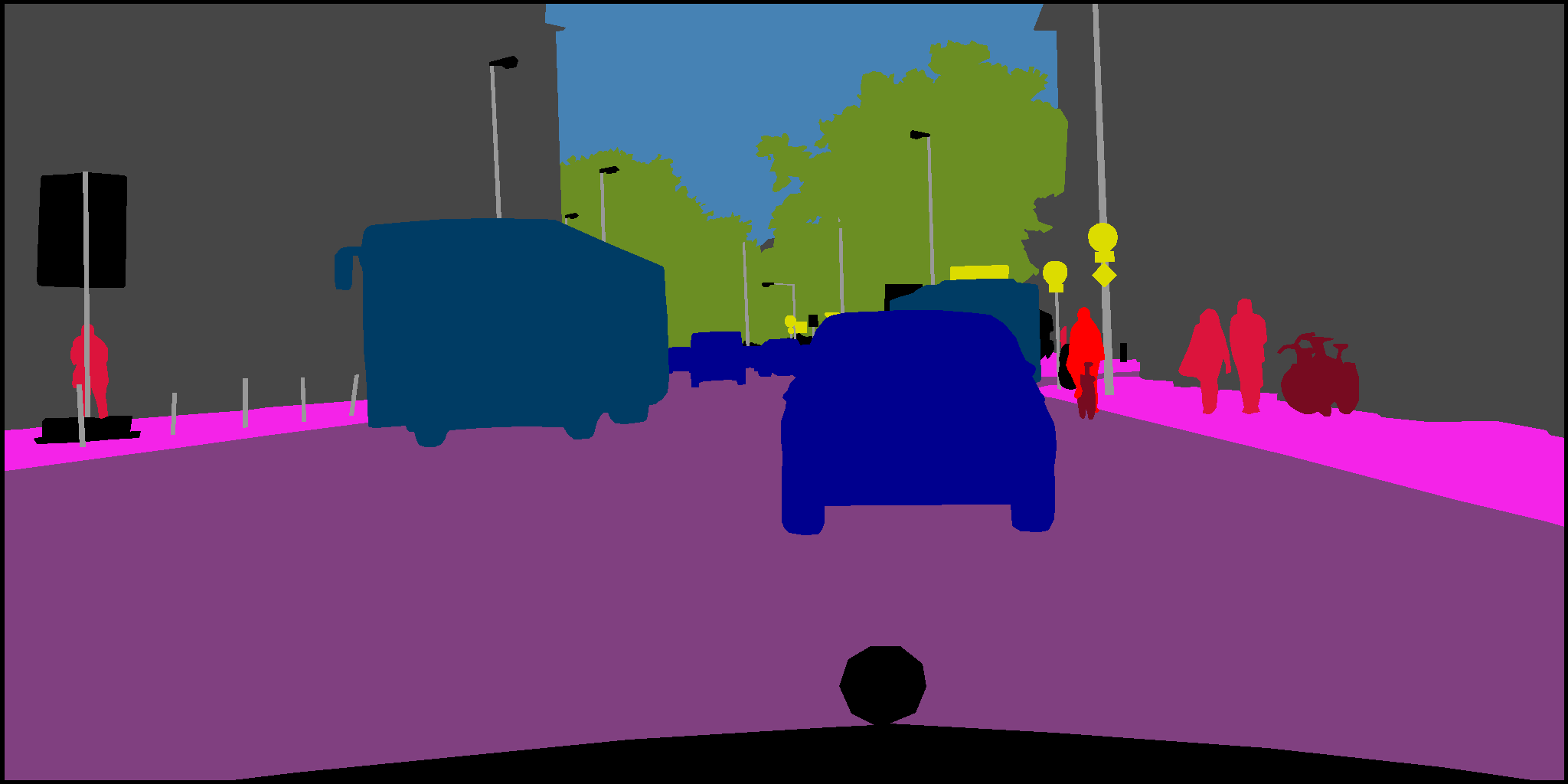} &
\includegraphics[width=\linewidth]{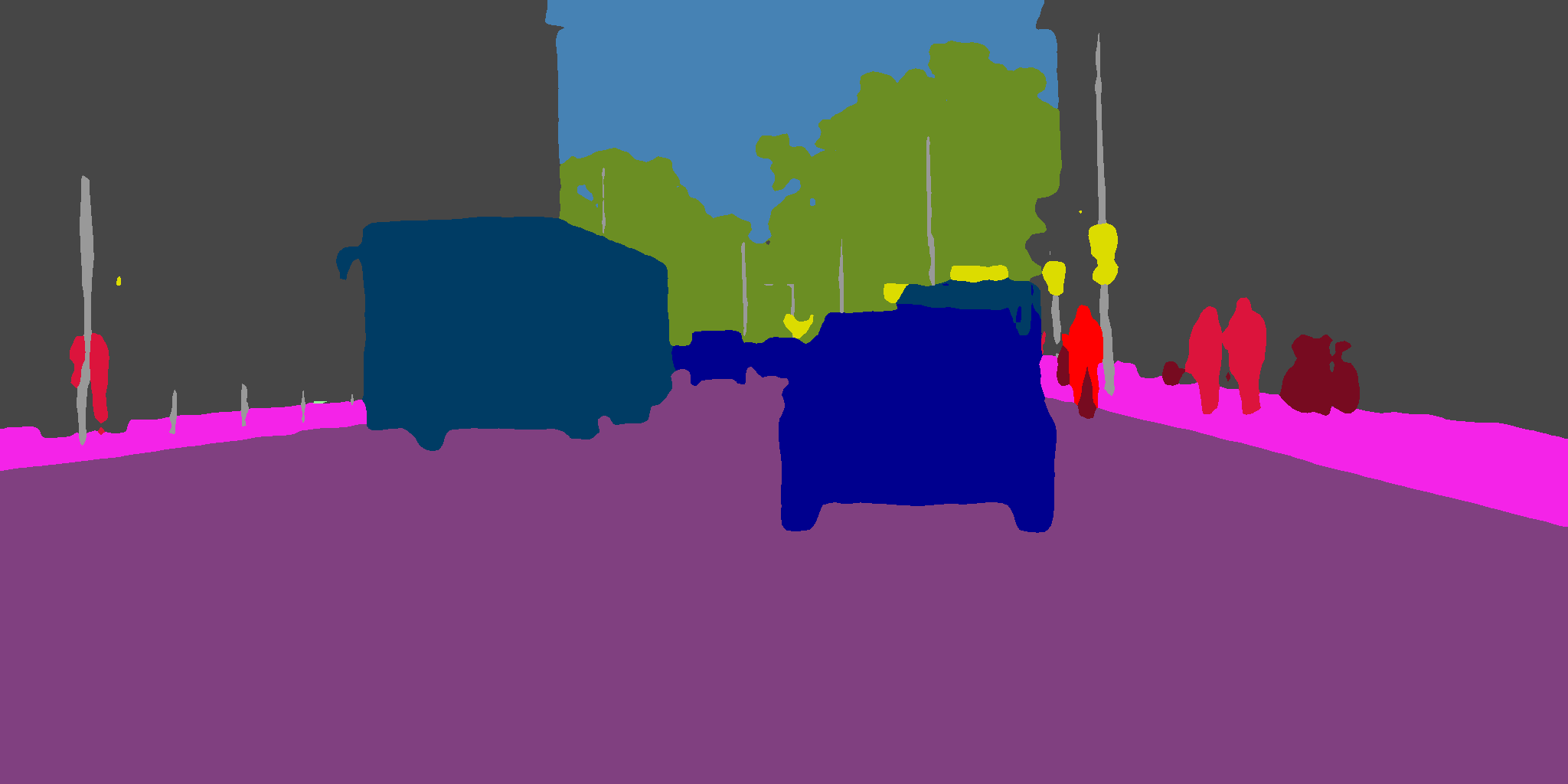} &
\includegraphics[width=\linewidth]{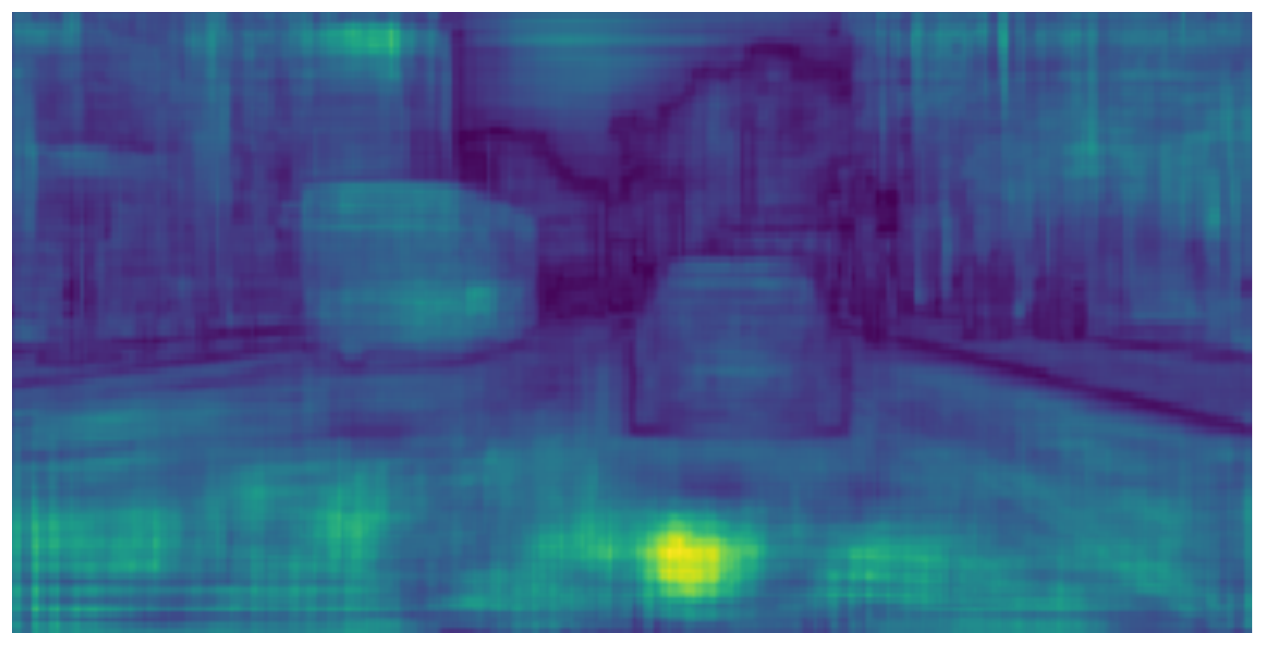} \\
\includegraphics[width=\linewidth]{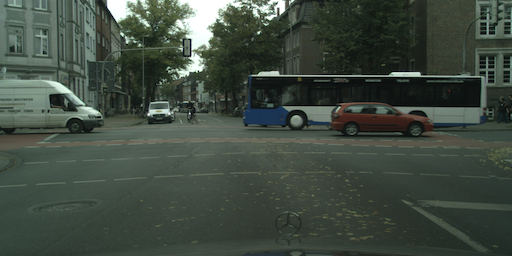} &
\includegraphics[width=\linewidth]{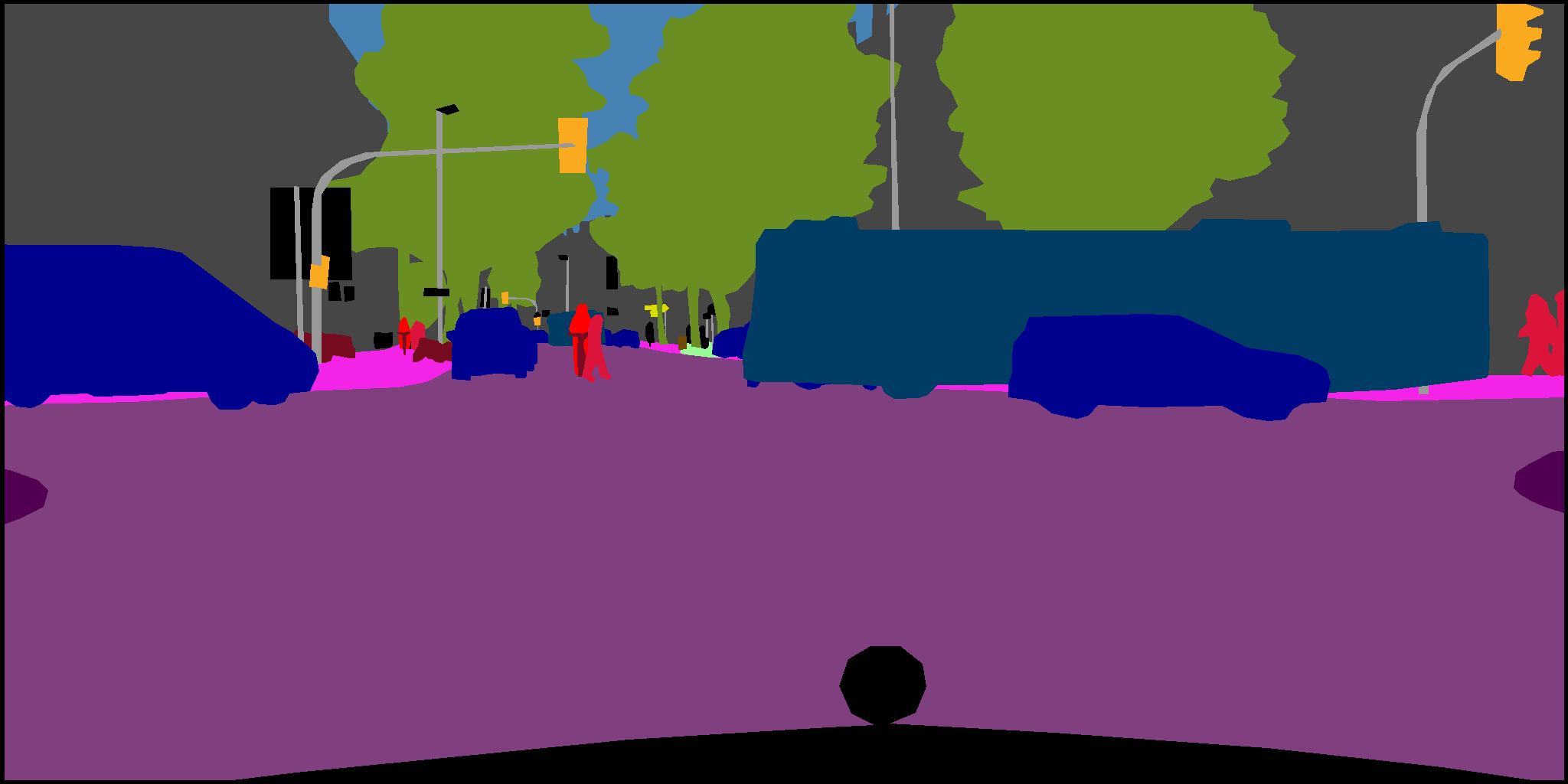} &
\includegraphics[width=\linewidth]{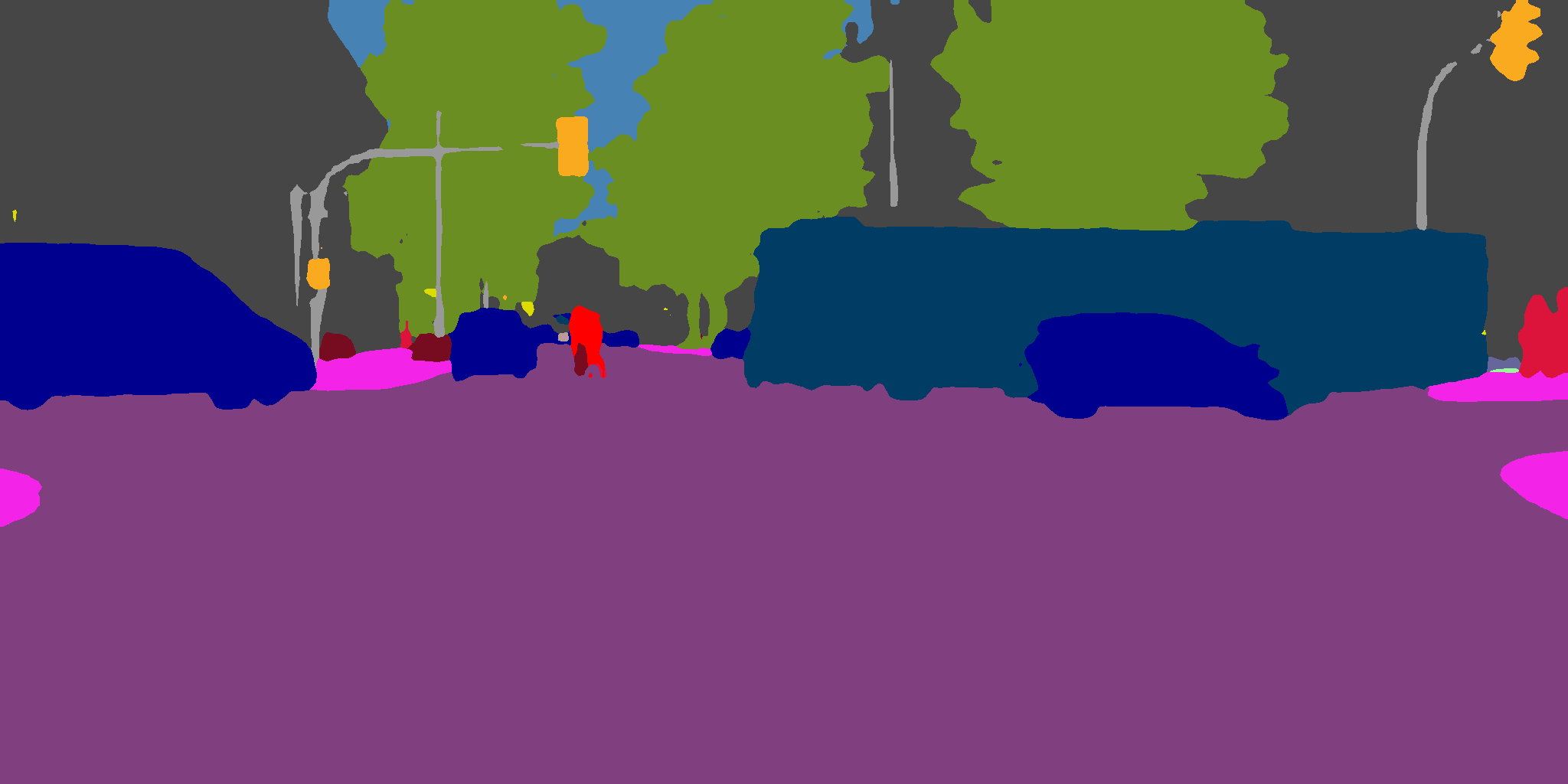} &
\includegraphics[width=\linewidth]{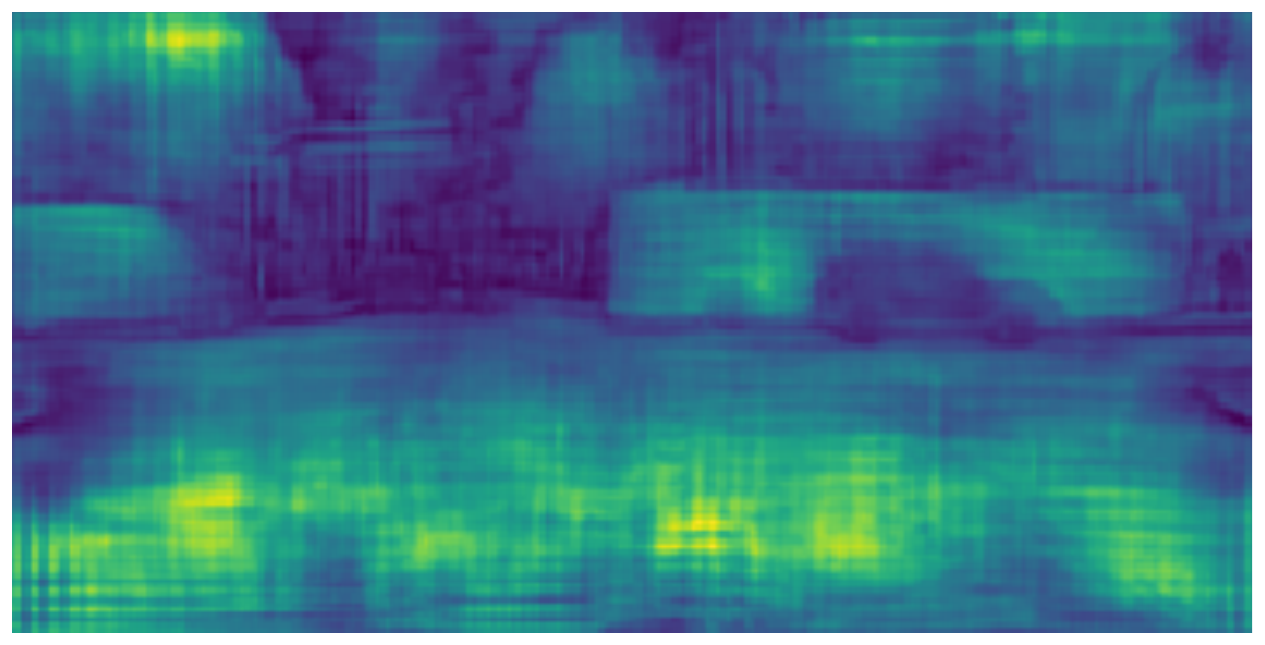} \\
\includegraphics[width=\linewidth]{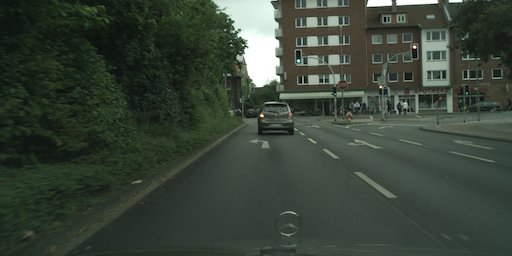} &
\includegraphics[width=\linewidth]{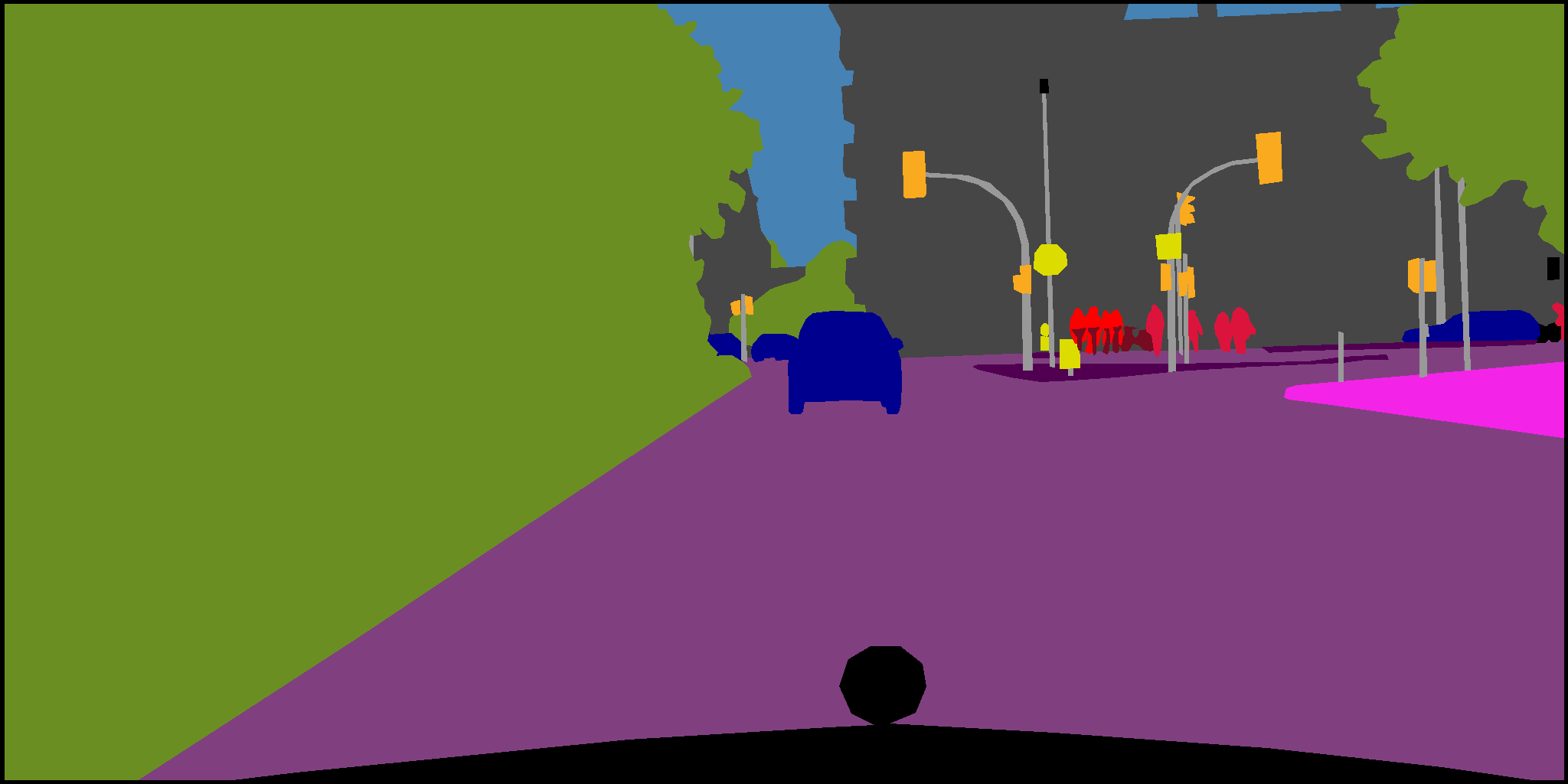} &
\includegraphics[width=\linewidth]{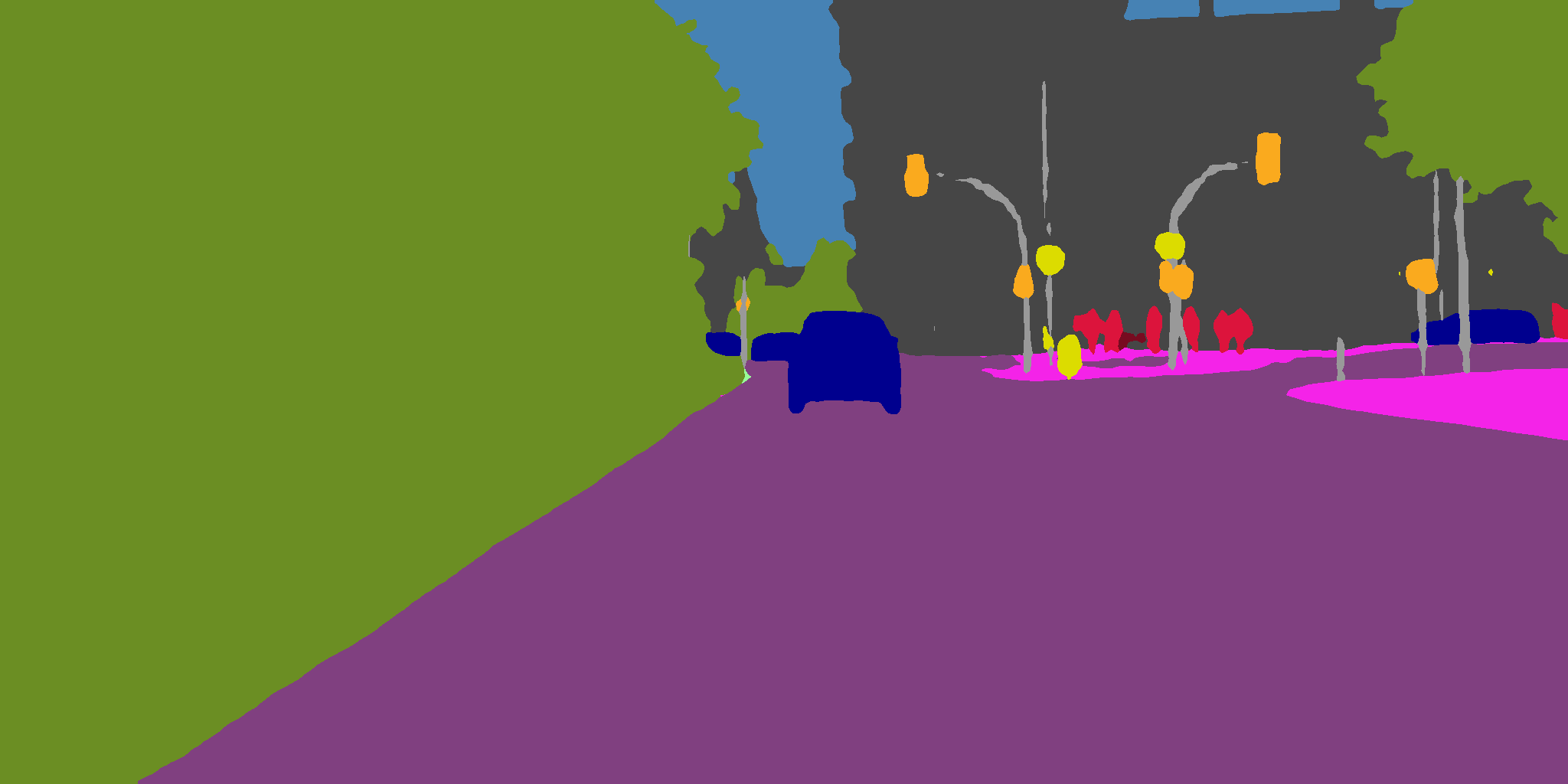} &
\includegraphics[width=\linewidth]{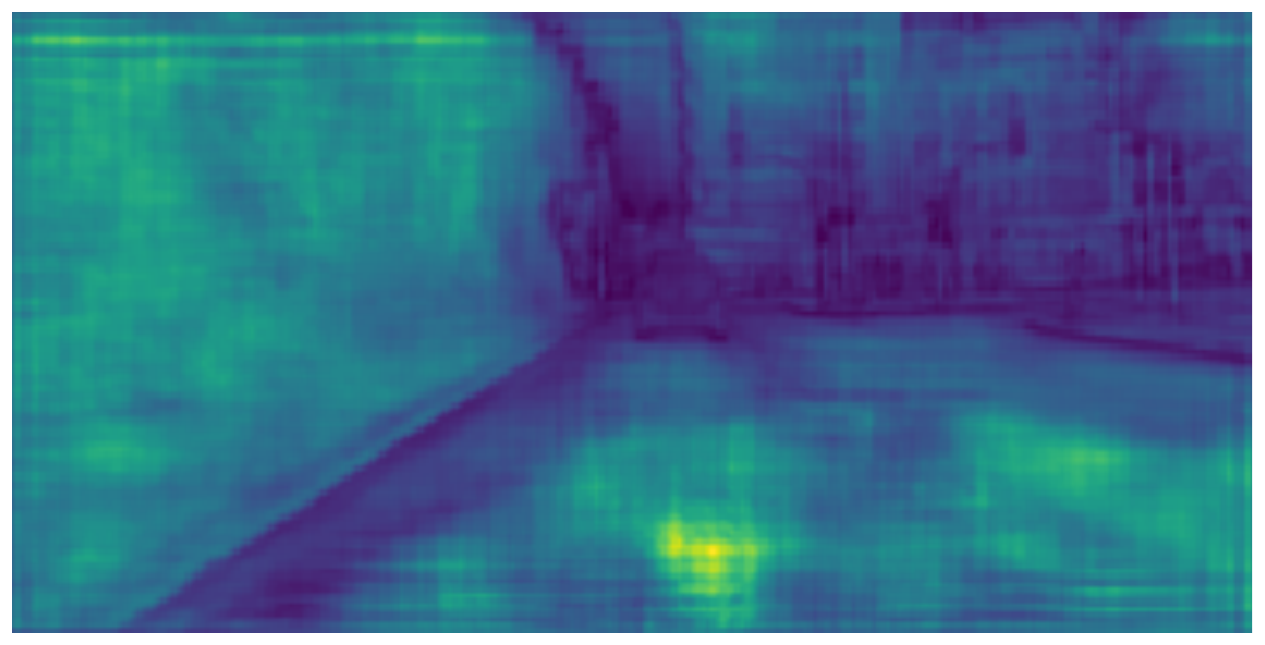} \\
\includegraphics[width=\linewidth]{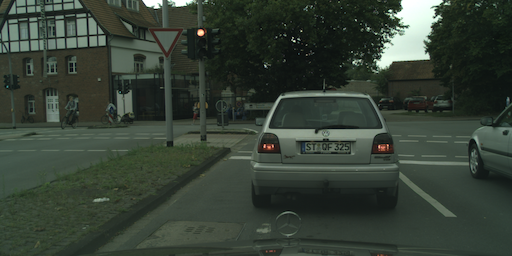} &
\includegraphics[width=\linewidth]{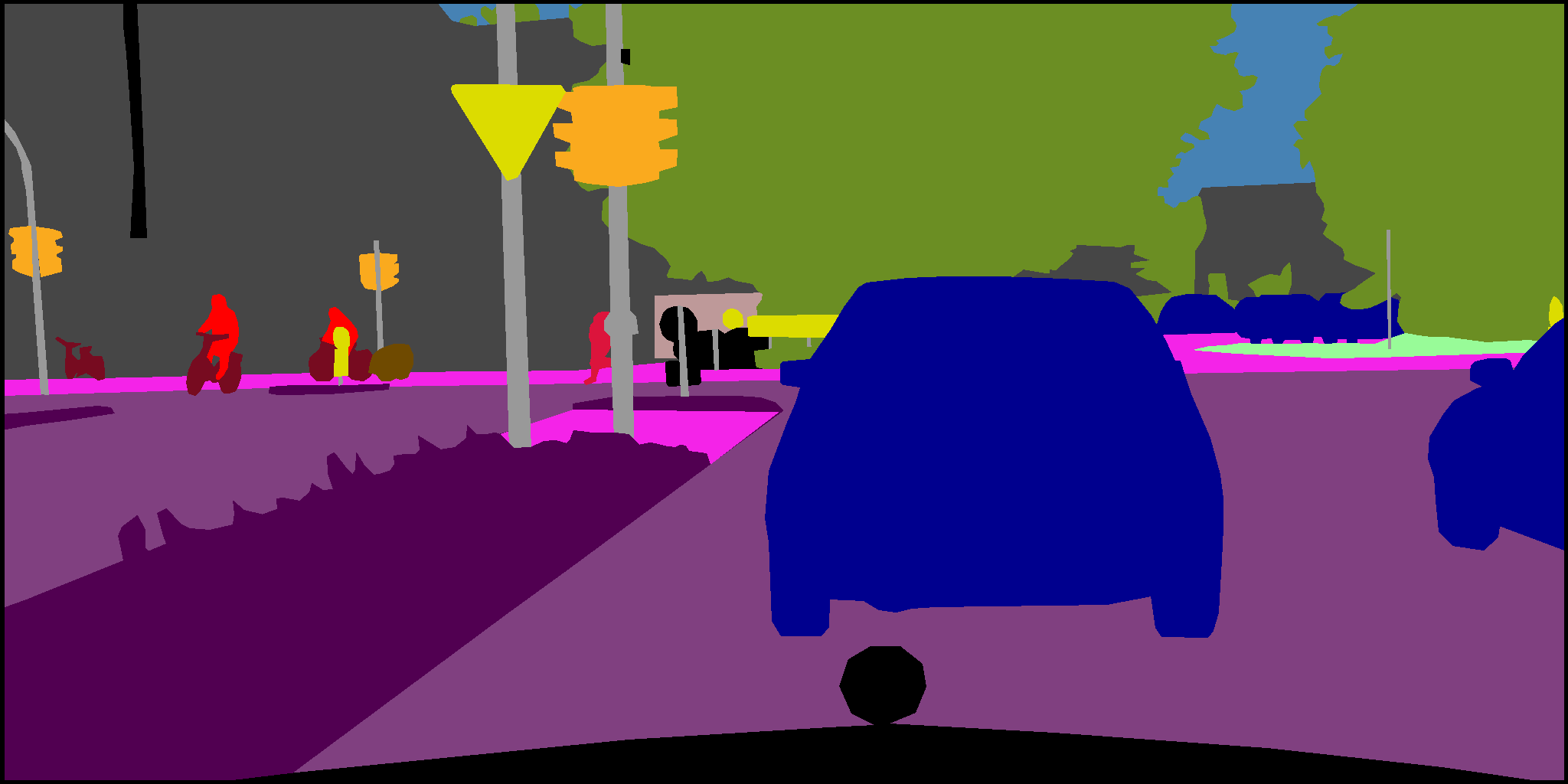} &
\includegraphics[width=\linewidth]{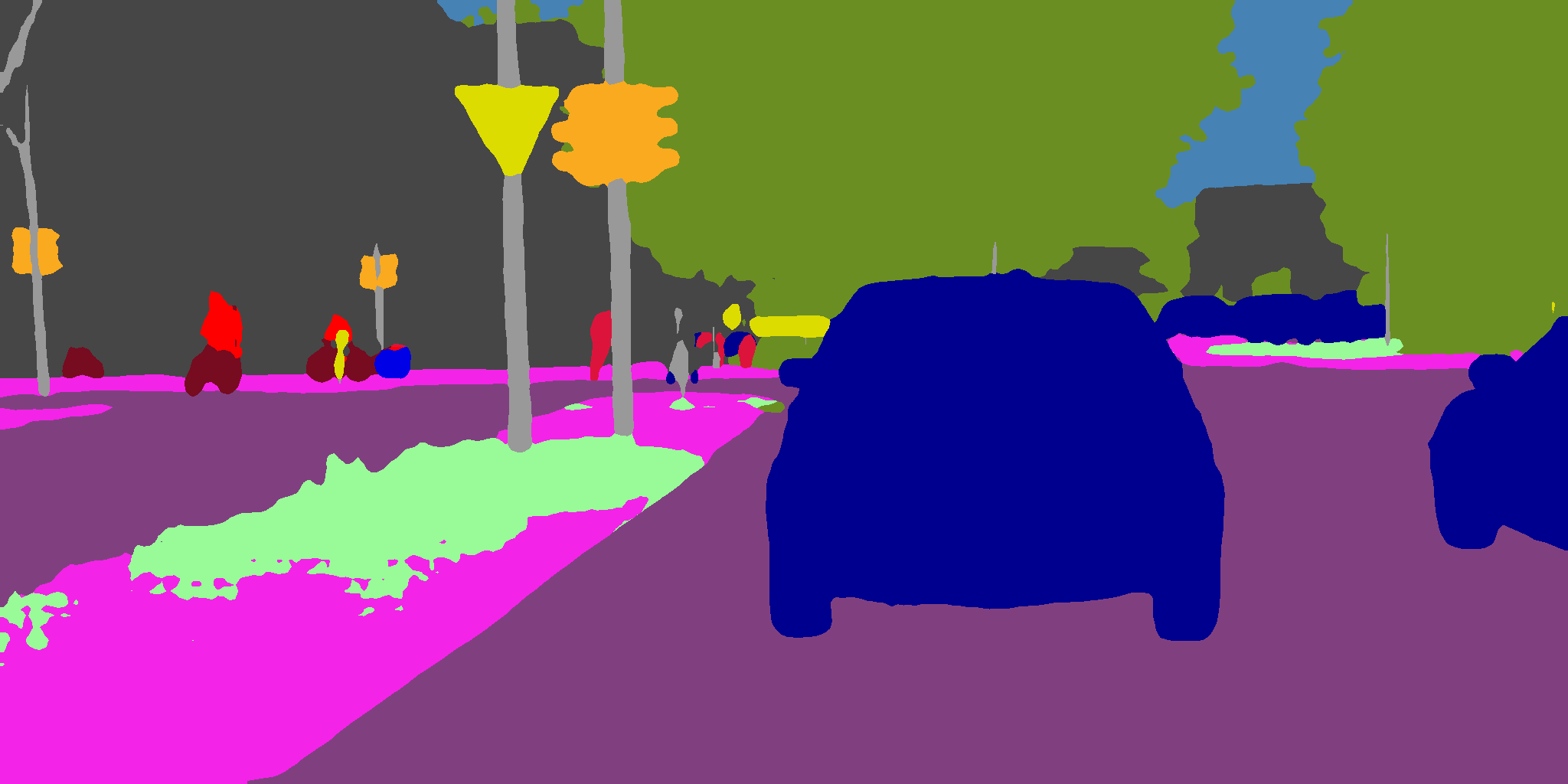} &
\includegraphics[width=\linewidth]{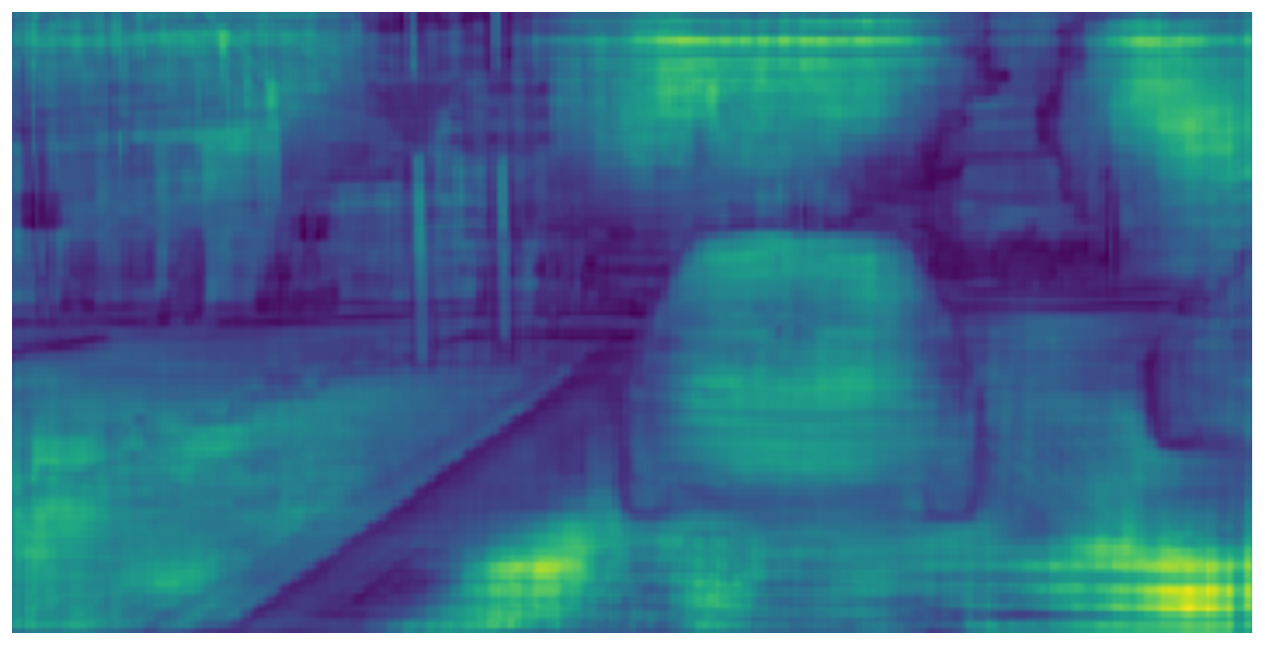} \\
\includegraphics[width=\linewidth]{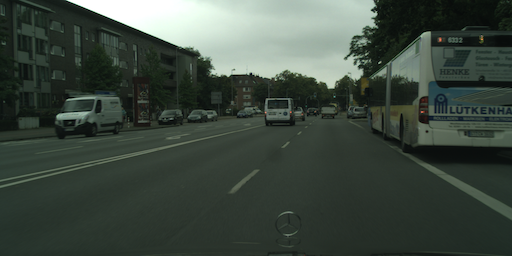} &
\includegraphics[width=\linewidth]{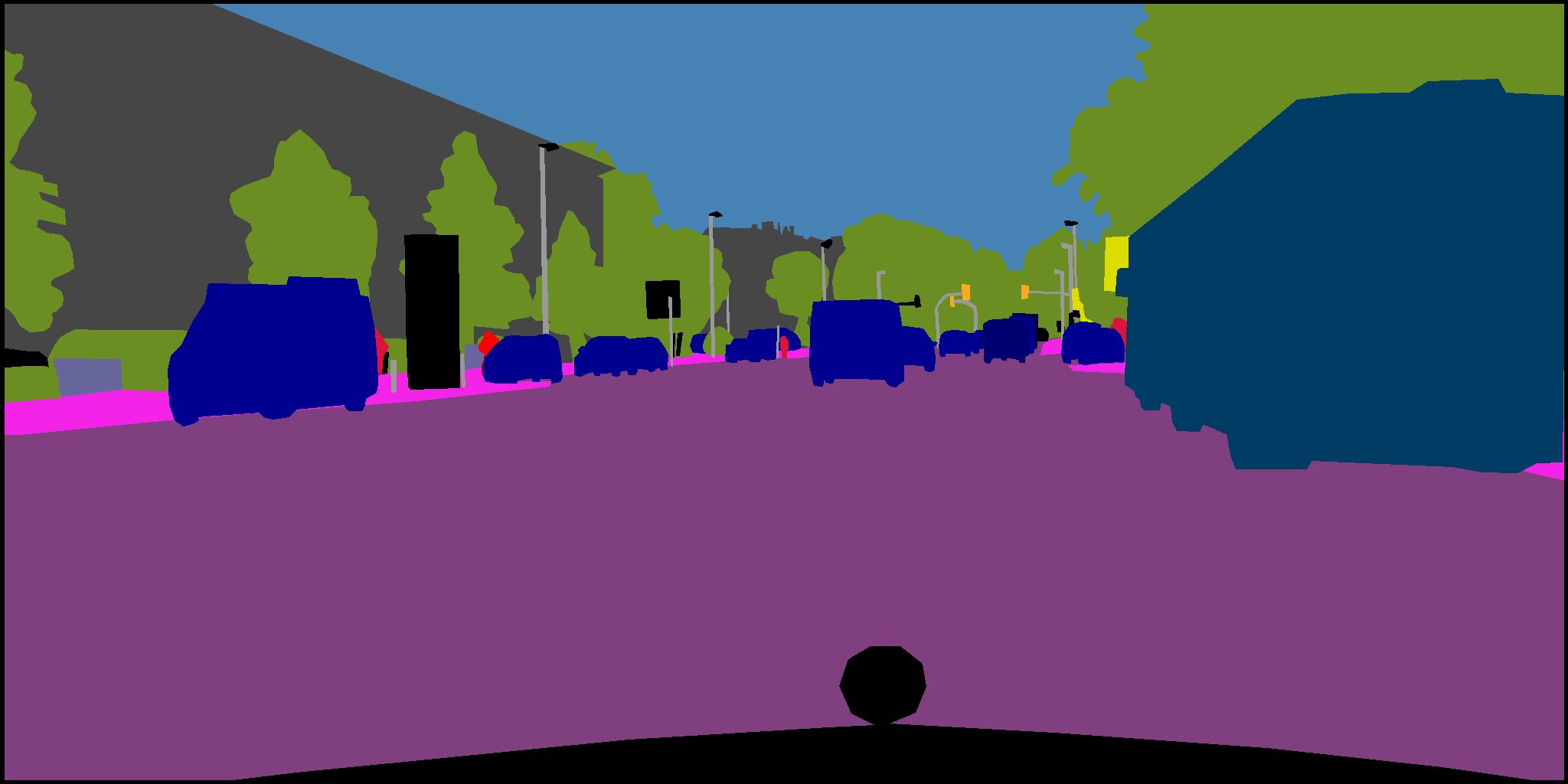} &
\includegraphics[width=\linewidth]{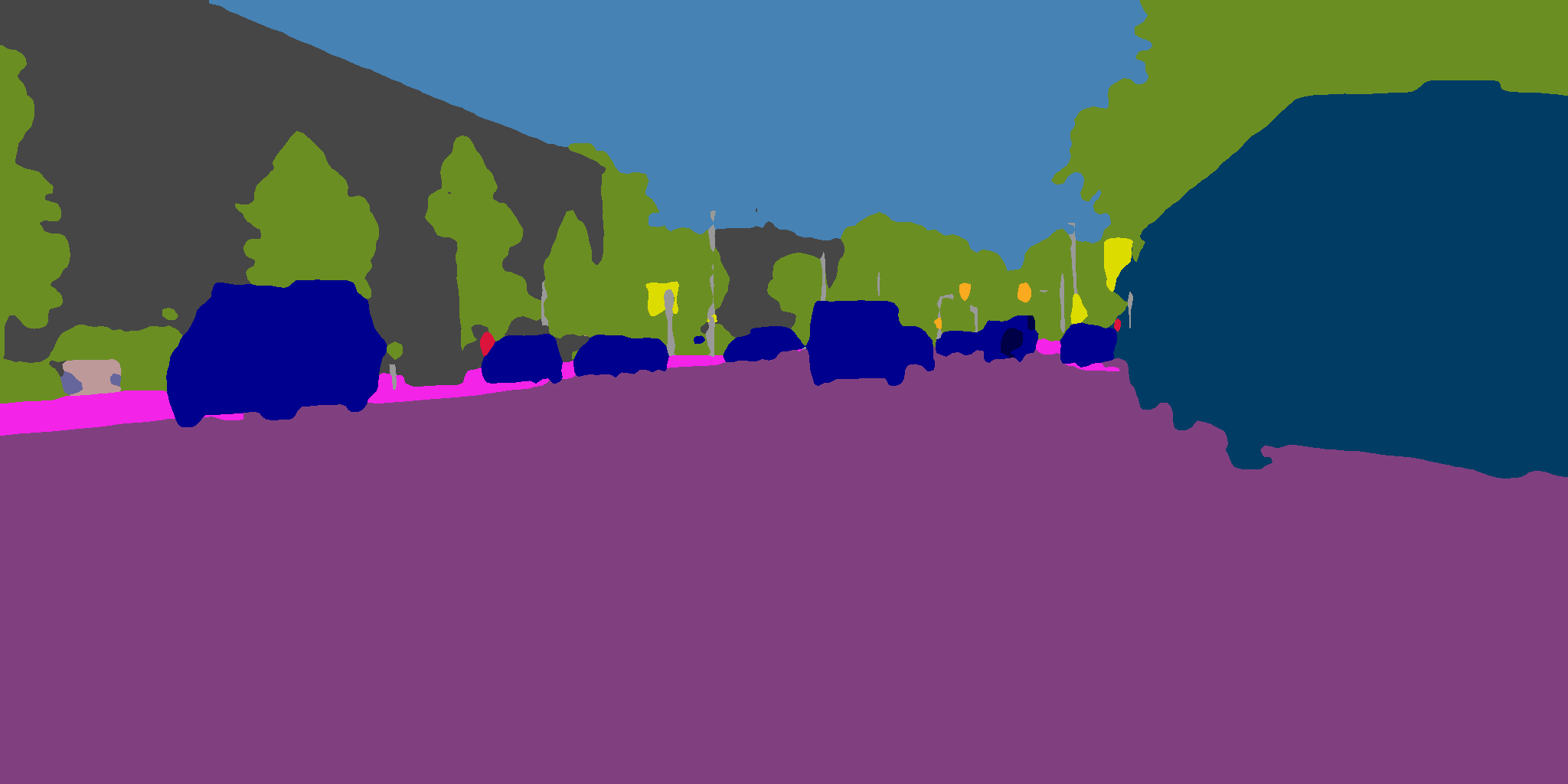} &
\includegraphics[width=\linewidth]{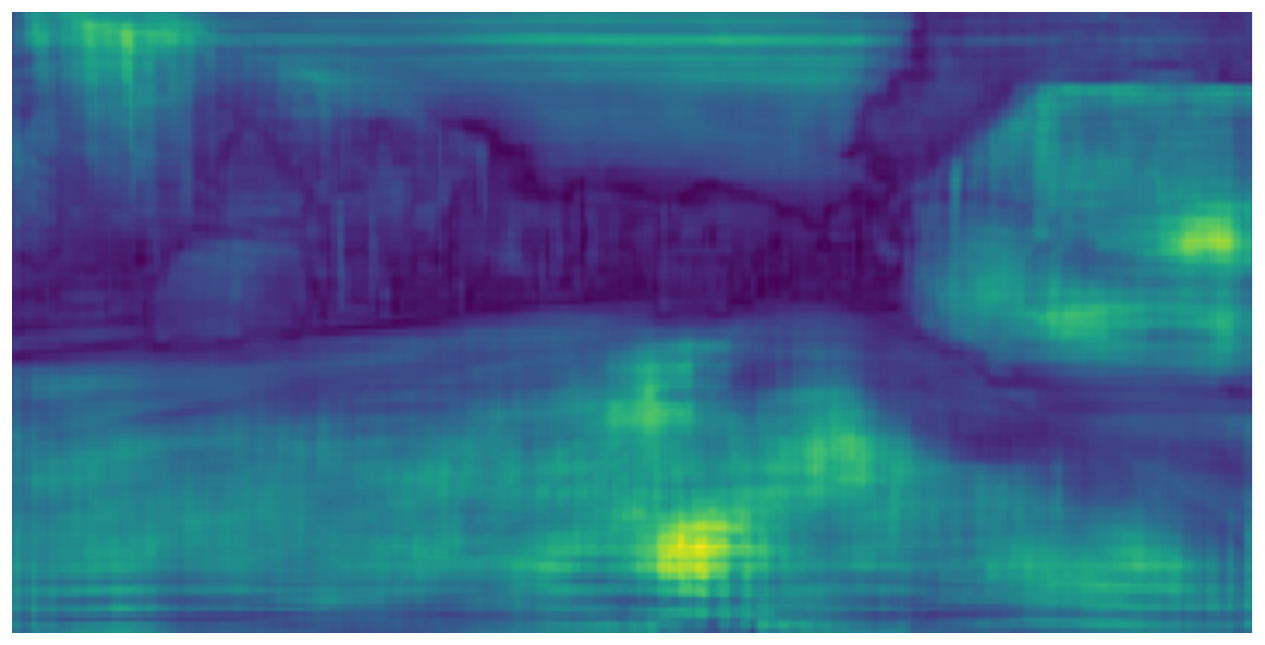} \\
\\
\resizebox{1in}{!}{(a)} & \resizebox{1in}{!}{(b)} & \resizebox{1in}{!}{(c)} & \resizebox{1in}{!}{(d)} \\
\end{tabular}
}
\end{center}
\caption{
Qualitative results for dynamic inference of scale: (a) input images; (b) truths; (c) outputs; and (d) scale estimates with small visualized as blue/dark and large visualized as yellow/bright.
The scale estimates have a certain amount of structure: coherent segments and boundaries between them can be seen.
}
\label{fig:sigma-visual}
\end{figure*}

Qualitative results for dynamic Gaussian structure are shown in Figure \ref{fig:sigma-visual}.
The inferred local scales reflect scale structure in the input.

In these experiments we restrict the Gaussian to spherical covariance with a single degree of freedom for scale.
Our results show that making scale dynamic through spherical covariance suffices to achieve essentially equal accuracy as general, free-form deformations.
Including further degrees of freedom by diagonal and full covariance does not give further improvement on this task and data.
As scale is perhaps the most ubiquitous transformation in the distribution of natural images, scale modeling might suffice to handle many variations.

\section{Conclusion}

Composing structured Gaussian and free-form filters makes receptive field size and shape differentiable for direct optimization.
Through receptive field learning, our semi-structured models do by gradient optimization what current free-form models have done by discrete design.
That is, in our parameterization changes in structured \emph{weights} would require changes in free-form \emph{architecture}.

Our method learns local receptive fields.
While we have focused on locality in space, the principle is more general, and extends to locality in time and other dimensions.

Factorization of this sort points to a reconciliation of structure and learning, through which known structure is respected and unknown detail is learned freely.

\begin{spacing}{1.0}
{\small
\bibliographystyle{ieee}
\bibliography{sigma}
}
\end{spacing}

\end{document}